\documentclass[10pt,journal,compsoc]{IEEEtran}
\usepackage{subfigure}
\usepackage{times}
\usepackage{url}

\usepackage[colorlinks,linkcolor=blue]{hyperref}
\usepackage{amsmath}
\usepackage{amssymb}
\usepackage{pifont}
\usepackage{array}
\usepackage{multirow}
\usepackage{lscape}
\usepackage{graphicx,epsfig,algorithmic,algorithm}
\usepackage{booktabs}
\usepackage{setspace}
\usepackage{subfigure}
\usepackage{epstopdf}
\usepackage{threeparttable}
\ifCLASSOPTIONcompsoc
\usepackage[nocompress]{cite}
\else
\usepackage{cite}
\fi
\ifCLASSINFOpdf
\else
\fi
\begin{document}
	
	%
	\title{Deep Facial Expression Recognition: A Survey}
	%
	%
	%
	%
	
	\author{Shan~Li 
		and~Weihong~Deng$^*$,~\IEEEmembership{Member,~IEEE}
		\IEEEcompsocitemizethanks{\IEEEcompsocthanksitem The authors are with the Pattern Recognition and Intelligent System Laboratory, School of Information and Communication Engineering, Beijing University of Posts and Telecommunications, Beijing, 100876, China. \protect\\
			E-mail:\{ls1995, whdeng\}@bupt.edu.cn.}
	}

	\IEEEtitleabstractindextext{%
		\begin{abstract}
			With the transition of facial expression recognition (FER) from laboratory-controlled to challenging in-the-wild conditions and the recent success of deep learning techniques in various fields, deep neural networks have  increasingly been leveraged to learn discriminative representations for automatic FER.
			Recent deep FER systems generally focus on two important issues: overfitting caused by a lack of sufficient training data and expression-unrelated variations, such as illumination, head pose and  identity bias.
			In this paper, we provide a comprehensive survey on deep FER, including datasets and algorithms that provide insights into these intrinsic problems.
			First, we introduce the available datasets that are widely used in the literature and provide accepted data selection and evaluation principles for these datasets. We then describe the standard pipeline of a deep FER system with the related background knowledge and suggestions of applicable implementations for each stage. For the state of the art in deep FER, we review existing novel deep neural networks and related training strategies that are designed for FER based on both static images and dynamic image sequences, and discuss their advantages and limitations. Competitive performances on widely used benchmarks are also summarized in this section. We then extend our survey to additional related issues and application scenarios. Finally, we review the remaining challenges and corresponding opportunities in this field as well as future directions for the design of robust deep FER systems.
		\end{abstract}
		
		}

	\maketitle

	\IEEEdisplaynontitleabstractindextext

	%
	\IEEEpeerreviewmaketitle

	\IEEEraisesectionheading{\section{Introduction}\label{sec:introduction}}

	%
	%
	%
	%
	\IEEEPARstart{F}{acial} expression is one of the most powerful, natural and universal signals for human beings to convey their emotional states and intentions \cite{darwin1998expression,tian2001recognizing}. 
	Numerous studies have been conducted on automatic facial expression analysis because of its practical importance in sociable robotics, medical treatment, driver fatigue surveillance, and many other human-computer interaction systems. 
	In the field of computer vision and machine learning, various facial expression recognition (FER) systems have been explored to encode expression information from facial representations.
	As early as the twentieth century, Ekman and Friesen \cite{ekman1971constants} defined six basic emotions based on cross-culture study \cite{ekman1994strong}, which indicated that humans perceive certain basic emotions in the same way regardless of culture. These prototypical facial expressions are anger, disgust, fear, happiness, sadness, and surprise. Contempt was subsequently added as one of the basic emotions \cite{matsumoto1992more}. Recently, advanced research on neuroscience and psychology argued that the model of six basic emotions are culture-specific and not universal \cite{jack2012facial}.
	
	Although the affect model based on basic emotions
	is limited in the ability to represent the complexity and subtlety of our daily affective displays \cite{zeng2009survey,sariyanidi2015automatic, martinez2016advances}, and other emotion description models, such as the Facial Action Coding System (FACS) \cite{ekman2002facial} and the continuous model using affect dimensions \cite{gunes2013categorical}, are considered to represent a wider range of emotions,
	the categorical model that describes emotions in terms of discrete  basic emotions is still the most popular perspective for FER, due to its pioneering investigations along with the direct and intuitive definition of facial expressions. And in this survey, we will limit our discussion on FER based on the categorical model.
	
	FER systems can be divided into two main categories according to the feature representations: static image FER and dynamic sequence FER. 
	In \textit{static-based methods} \cite{shan2009facial, liu2014facial, mollahosseini2016going}, the feature representation is encoded with only spatial information from the current single image, whereas \textit{dynamic-based methods} \cite{zhao2007dynamic, jung2015joint, zhao2016peak} consider the temporal relation among contiguous frames in the input facial expression sequence. Based on these two vision-based methods, other modalities, such as audio and physiological channels, have also been used in \textit{multimodal systems} \cite{corneanu2016survey} to assist the recognition of expression.
	
\begin{figure}
		\centering
		\includegraphics[width=8.5cm]{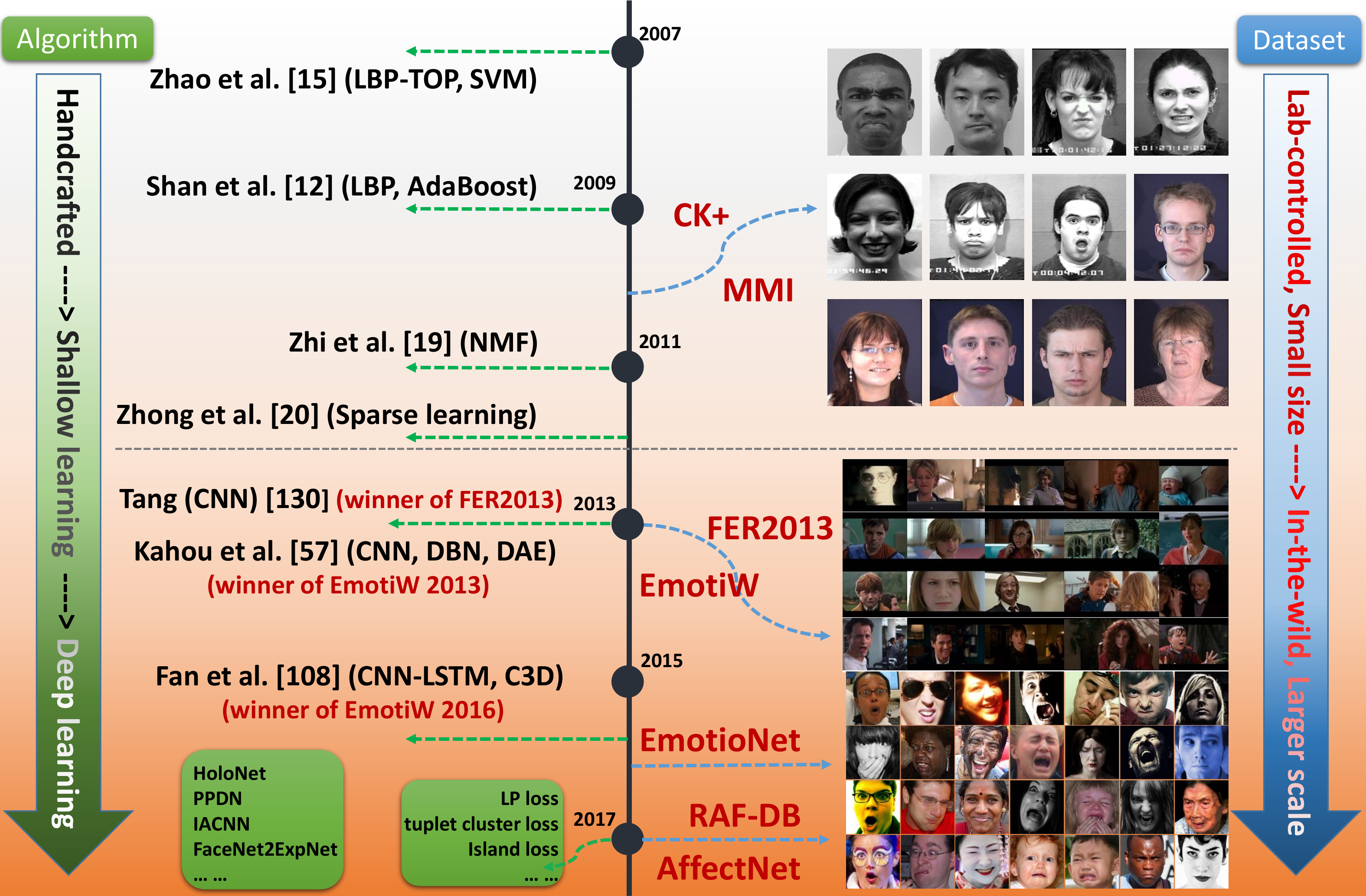}
		\caption{The evolution of facial expression recognition in terms of datasets and methods.}
		\label{basic}
	\end{figure}

	The majority of the traditional methods have used handcrafted features or shallow learning (e.g., local binary patterns (LBP) \cite{shan2009facial}, LBP on three orthogonal planes (LBP-TOP) \cite{zhao2007dynamic}, non-negative matrix factorization (NMF) \cite{zhi2011graph} and sparse learning \cite{zhong2012learning}) for FER. However, since 2013, emotion recognition competitions such as FER2013 \cite{goodfellow2013challenges} and Emotion Recognition in the Wild (EmotiW) \cite{dhall2015video,dhall2016emotiw,dhall2017individual} have collected  relatively sufficient training data from challenging real-world scenarios, which implicitly promote the transition of FER from lab-controlled to in-the-wild settings. 
	In the meanwhile, due to the dramatically increased chip processing abilities (e.g., GPU units) and well-designed network architecture, studies in various fields have begun to transfer to deep learning methods, which have achieved the state-of-the-art recognition accuracy and exceeded previous results by a large margin (e.g., \cite{krizhevsky2012imagenet,simonyan2014very,szegedy2015going,he2016deep}). Likewise, given with more effective training data of facial expression, deep learning techniques have increasingly been implemented to handle the challenging factors for emotion recognition in the wild. Figure \ref{basic} illustrates this evolution on FER in the aspect of algorithms and datasets.
	
	Exhaustive surveys on automatic expression analysis have been published in recent years  \cite{pantic2000automatic,fasel2003automatic,zeng2009survey,sariyanidi2015automatic}. These surveys have established a set of standard algorithmic pipelines for FER. However, they focus on traditional methods, and deep learning has rarely been reviewed. Very recently, FER based on deep learning has been surveyed in \cite{zhang2017survey}, which is a brief review without introductions on FER datasets and  technical details on deep FER. Therefore, in this paper, we make a systematic  research on deep learning for FER tasks based on both static images and videos (image sequences). We aim to give a newcomer to this filed an overview of the systematic framework and prime skills for deep FER.
	
	Despite the powerful feature learning ability of deep learning, problems remain when applied to FER. First, deep neural networks require a large amount of training data to avoid overfitting. However, the existing facial expression databases are not sufficient to train the well-known neural network with deep architecture that achieved the most promising results in object recognition tasks. Additionally,  high inter-subject variations exist due to different personal attributes, such as age, gender, ethnic backgrounds and level of expressiveness \cite{valstar2012meta}. In addition to subject identity bias, variations in pose, illumination and occlusions are common in unconstrained facial expression scenarios. These factors are nonlinearly coupled with facial expressions and therefore strengthen the requirement of deep networks to address the large intra-class variability and to learn effective expression-specific  representations.
	
	In this paper, we introduce recent advances in research on solving the above problems for deep FER. We examine the state-of-the-art results that have not been reviewed in previous survey papers. The rest of this paper is organized as follows. Frequently used expression databases are introduced in Section \ref{database}. Section \ref{FER} identifies three main steps required in a deep FER system and describes the related background.  Section \ref{method} provides a detailed review of novel neural network architectures and special network training tricks designed for FER based on static images and dynamic image sequences. We then cover additional related issues and other practical scenarios in Section \ref{addition}. Section \ref{further} discusses some of the challenges and opportunities in this field and identifies potential future directions.
	\section{Facial expression databases}
	\label{database}
	Having sufficient labeled training data that include as many variations of the populations and environments  as possible is important for the design of a deep expression recognition system. In this section, we discuss the publicly available databases that contain basic expressions and that are widely used in our reviewed papers for deep learning algorithms evaluation. 
	We also introduce newly released databases that contain a large amount of affective images collected from the real world to benefit the training of deep neural networks. Table \ref{databases} provides an overview of these datasets, including the main reference, number of subjects, number of image or video samples, collection environment, expression distribution and additional information.
	\begin{table*}[ht]
		\setlength{\tabcolsep}{7pt}
		\centering
		\renewcommand{\arraystretch}{1.5}
		\scriptsize
		\caption{An overview of the facial expression datasets. P = posed; S = spontaneous; Condit. = Collection condition; Elicit. = Elicitation method.}   \label{databases}
		\begin{tabular}{m{0.13\textwidth}<{\centering}||m{0.08\textwidth}<{\centering}|c|c|c|m{0.2\textwidth}<{\centering}|m{0.3\textwidth}<{\centering}}
			\hline
			\textbf{Database} & \textbf{Samples} & \textbf{Subject} & \textbf{Condit.} & \textbf{Elicit.} & \textbf{Expression distribution} & \textbf{Access} \\\hline\hline
			CK+~\cite{lucey2010extended} & 593 image sequences & 123 & Lab  & P \& S & 6 basic expressions plus contempt and neutral & \url{http://www.consortium.ri.cmu.edu/ckagree/} \\\hline
			
			MMI~\cite{pantic2005web,valstar2010induced} & 740 images and 2,900 videos   & 25 & Lab   & P & 6 basic expressions plus neutral & \url{https://mmifacedb.eu/} \\\hline
			
			JAFFE~\cite{lyons1998japanese} & 213 images   & 10  & Lab      & P & 6 basic expressions plus neutral  & \url{http://www.kasrl.org/jaffe.html} \\\hline
			
			TFD~\cite{susskind2010toronto}&112,234 images&N/A&Lab &P&6 basic expressions plus neutral &\url{josh@mplab.ucsd.edu}\\\hline
			
			FER-2013~\cite{goodfellow2013challenges}&35,887 images&N/A&Web&P \& S&6 basic expressions plus neutral&\url{https://www.kaggle.com/c/challenges-in-representation-learning-facial-expression-recognition-challenge}\\\hline
			
			AFEW 7.0~\cite{dhall2017individual} & 1,809 videos   & N/A  & Movie    & P \& S & 6 basic expressions plus neutral &\url{https://sites.google.com/site/emotiwchallenge/} \\\hline
			
			SFEW 2.0~\cite{dhall2015video} & 1,766 images & N/A  & Movie  & P \& S & 6 basic expressions plus neutral &\url{https://cs.anu.edu.au/few/emotiw2015.html}\\\hline
			
			Multi-PIE~\cite{gross2010multi} &  755,370 images & 337 & Lab     & P & Smile, surprised, squint, disgust, scream and neutral & \url{http://www.flintbox.com/public/project/4742/} \\\hline
			
			BU-3DFE~\cite{yin20063d} & 2,500 images & 100 & Lab     & P & 6 basic expressions plus neutral  & \url{http://www.cs.binghamton.edu/~lijun/Research/3DFE/3DFE_Analysis.html} \\\hline
			
			Oulu-CASIA \cite{zhao2011facial}&2,880 image sequences&80&Lab&P&6 basic expressions&\url{http://www.cse.oulu.fi/CMV/Downloads/Oulu-CASIA}\\\hline
			
			RaFD~\cite{langner2010presentation} & 1,608 images  & 67   & Lab      & P& 6 basic expressions plus contempt and neutral & \url{http://www.socsci.ru.nl:8180/RaFD2/RaFD}\\\hline
			
			KDEF \cite{lundqvist1998karolinska}&4,900 images&70&Lab&P &6 basic expressions plus neutral&\url{http://www.emotionlab.se/kdef/}\\\hline
			
			EmotioNet~\cite{benitez2016emotionet}&1,000,000 images&N/A&Web& P \& S &23 basic expressions or compound expressions&\url{http://cbcsl.ece.ohio-state.edu/dbform_emotionet.html}\\\hline
			
			RAF-DB \cite{li2017reliable,li2018reliable}& 29672 images & N/A & Web  & P \& S & 6 basic expressions plus neutral and 12 compound expressions & \url{http://www.whdeng.cn/RAF/model1.html} \\\hline
			
			AffectNet~\cite{Mollahosseini2017AffectNet}&450,000 images (labeled)&N/A&Web& P \& S &6 basic expressions plus neutral&\url{http:
				//mohammadmahoor.com/databases-codes/}\\\hline
			
			ExpW \cite{zhang2018From}&91,793 images&N/A&Web&P \& S&6 basic expressions plus neutral&\url{http://mmlab.ie.cuhk.edu.hk/projects/socialrelation/index.html}\\\hline
			
			\hline
		\end{tabular}%
	\end{table*}%
	
	\textbf{CK+ \cite{lucey2010extended}:} The Extended Cohn–Kanade (CK+) database is the most extensively used laboratory-controlled database for evaluating FER systems. CK+ contains 593 video sequences from 123 subjects. The sequences vary in duration from 10 to 60 frames and show a shift from a neutral facial expression to the peak expression. Among these videos, 327 sequences from 118 subjects are labeled with seven basic expression labels (anger, contempt, disgust, fear, happiness, sadness, and surprise) based on the Facial Action Coding System (FACS). Because CK+ does not provide specified training, validation and test sets, the algorithms evaluated on this database are not uniform. For static-based methods, the most common data selection method is to extract the last one to three frames with peak formation and the first frame (neutral face) of each sequence. Then, the subjects are divided  into $n$ groups for person-independent $n$-fold cross-validation experiments, where commonly selected  values of $n$ are 5, 8 and 10.
	
	\textbf{MMI~\cite{pantic2005web,valstar2010induced}:} The MMI database is  laboratory-controlled and includes 326 sequences from 32 subjects. A total of 213 sequences are labeled with six basic expressions (without ``contempt''), and 205 sequences are captured in frontal view. In contrast to CK+, sequences in MMI are onset-apex-offset  labeled, i.e., the sequence begins with a neutral expression and reaches peak
	near the middle before returning to the neutral expression. Furthermore, MMI has more challenging conditions, i.e., there are large inter-personal variations because subjects perform the same expression non-uniformly and many of them wear accessories (e.g., glasses, mustache). For experiments, the most common method is to choose the first frame (neutral face) and the three peak frames in each frontal sequence to conduct person-independent 10-fold cross-validation.
	
	\textbf{JAFFE~\cite{lyons1998japanese}:} The Japanese Female Facial Expression (JAFFE) database is a laboratory-controlled image database that contains 213 samples of posed expressions from 10 Japanese females. Each person has 3\~{}4 images with each of six basic facial expressions (anger, disgust, fear, happiness, sadness, and surprise) and one image with a neutral expression. The database is challenging because it contains few examples per subject/expression. Typically, all the images are used for the leave-one-subject-out experiment.
	
	\textbf{TFD \cite{susskind2010toronto}:}The Toronto Face Database (TFD) is an amalgamation of several facial expression datasets. TFD contains 112,234 images, 4,178 of which are annotated with one of seven expression labels: anger, disgust, fear, happiness, sadness, surprise and neutral. The faces have already been detected and normalized to a size of 48*48 such that all the subjects’ eyes are the same distance apart and have the same vertical coordinates.
	Five official folds are provided in TFD; each fold contains a training, validation, and test set consisting of 70\%, 10\%, and 20\% of the images, respectively.
	
	\textbf{FER2013~\cite{goodfellow2013challenges}:} The FER2013 database was introduced during the ICML 2013 Challenges in Representation Learning. FER2013 is a large-scale and unconstrained database collected automatically by the Google image search API. All images have been registered and resized to 48*48 pixels after rejecting wrongfully labeled frames and adjusting the cropped region. FER2013 contains  28,709 training images, 3,589 validation images and 3,589 test images with seven expression labels (anger, disgust, fear, happiness, sadness,  surprise and neutral).
	
	\textbf{AFEW \cite{dhall2012collecting}:} The Acted Facial Expressions in the Wild (AFEW) database was first established and introduced in \cite{dhall2011acted}  and has served as an evaluation platform for the annual Emotion Recognition In The Wild Challenge (EmotiW) since 2013. AFEW contains video clips collected from different movies with spontaneous expressions, various head poses, occlusions and illuminations. AFEW is a temporal and multimodal database that provides with vastly different environmental conditions in both audio and video. Samples are labeled with seven expressions: anger, disgust, fear, happiness, sadness, surprise and neutral. The annotation of expressions have been continuously updated, and reality TV show data have been continuously added. The \textbf{AFEW 7.0} in EmotiW 2017 \cite{dhall2017individual} is divided into three data partitions in an independent manner in terms of subject and movie/TV source: Train (773 samples), Val (383 samples) and Test (653 samples), 
	which ensures data in the three sets belong to mutually exclusive movies and actors.
	

	\textbf{SFEW~\cite{dhall2011static}:} The Static Facial Expressions in the Wild (SFEW) was created by selecting static frames from the AFEW database by computing key frames based on facial point clustering. The most commonly used version, \textbf{SFEW 2.0}, was the benchmarking data for the SReco sub-challenge in EmotiW 2015 \cite{dhall2015video}. SFEW 2.0 has been divided into three sets: Train (958 samples), Val (436 samples) and Test (372 samples). Each of the images is assigned to one of seven expression categories, i.e., anger, disgust, fear, neutral, happiness, sadness, and surprise. The expression labels of the training and validation sets are publicly available, whereas those of the testing set are held back by the challenge organizer.
	
	\textbf{Multi-PIE \cite{gross2010multi}:} The CMU Multi-PIE database contains 755,370 images from 337 subjects under 15 viewpoints and 19 illumination conditions in up to four recording session. Each facial image is labeled with one of six expressions: disgust, neutral, scream, smile, squint and surprise. This dataset is typically used for multiview facial expression analysis.
	
	\textbf{BU-3DFE \cite{yin20063d}:} The Binghamton University 3D Facial Expression (BU-3DFE)  database  contains 606 facial expression sequences captured from 100 people. For each subject, six universal facial expressions (anger, disgust, fear, happiness, sadness and surprise) are elicited by various manners with multiple intensities. Similar to Multi-PIE, this dataset is typically used for multiview 3D facial expression analysis.
	
	\textbf{Oulu-CASIA \cite{zhao2011facial}:} The Oulu-CASIA database includes 2,880 image sequences collected from 80 subjects labeled with six basic emotion labels: anger, disgust, fear, happiness, sadness, and surprise.  Each of the videos is captured with one of two imaging systems, i.e., near-infrared (NIR) or visible light (VIS), under three different illumination conditions. Similar to CK+, the first frame is neutral and the last frame has the peak expression. Typically, only the last three peak frames and the first frame (neutral face) from the 480 videos collected by the VIS System under normal indoor illumination are employed for 10-fold cross-validation experiments.
	
	\textbf{RaFD \cite{langner2010presentation}:} The Radboud Faces Database (RaFD) is laboratory-controlled and has a total of 1,608 images from 67 subjects with three different gaze directions, i.e., front, left and right. Each sample is labeled with one of eight expressions: anger, contempt, disgust, fear, happiness, sadness, surprise and neutral.
	
	\textbf{KDEF \cite{lundqvist1998karolinska}:} The laboratory-controlled Karolinska Directed Emotional Faces (KDEF) database was originally developed for use in psychological and medical research. KDEF consists of images from 70 actors with five different angles labeled with six basic facial expressions plus neutral.
	
	In addition to these commonly used datasets for basic emotion recognition, several well-established and large-scale publicly available facial expression databases collected from the Internet that are suitable for training deep neural networks have emerged in the last two years. 
	
	\textbf{EmotioNet~\cite{benitez2016emotionet}:} EmotioNet is a large-scale database with one million facial expression images collected from the Internet. A total of 950,000 images were annotated by the automatic action unit (AU) detection model in \cite{benitez2016emotionet}, and the remaining 25,000 images were manually annotated with 11 AUs. The second track of the EmotioNet Challenge \cite{benitez2017emotionet} provides six basic expressions and ten compound expressions \cite{du2014compound}, and 2,478 images with expression labels are available.
	
	\textbf{RAF-DB~\cite{li2017reliable,li2018reliable}:} The Real-world Affective Face Database (RAF-DB) is a real-world database that contains 29,672 highly diverse facial images downloaded from the Internet. With manually crowd-sourced annotation and reliable estimation, seven basic and eleven compound emotion labels are provided for the samples. 
	Specifically, 15,339 images from the basic emotion set are divided into two groups (12,271 training samples and 3,068 testing samples) for evaluation.
	
	\textbf{AffectNet~\cite{Mollahosseini2017AffectNet}:} AffectNet contains more than one million images from the Internet that were obtained by querying different search engines using emotion-related tags. It is by far the largest database that provides facial expressions in two different emotion models (categorical model and dimensional model), of which 450,000 images have manually annotated labels for eight basic expressions.
	
	\textbf{ExpW \cite{zhang2018From}:} The Expression in-the-Wild Database (ExpW) contains 91,793 faces downloaded using Google image search. Each of the face images was manually annotated as one of the seven basic expression categories. Non-face images were removed in the annotation process.
	
	\section{Deep facial expression recognition}
	\label{FER}
	In this section, we describe the three main steps that are common in automatic deep FER, i.e., pre-processing, deep feature learning and deep feature classification. We briefly summarize the widely used algorithms for each step and recommend the existing state-of-the-art best practice implementations according to the referenced papers.
	\begin{figure*}[t]
		\centering
		\includegraphics[width=18cm]{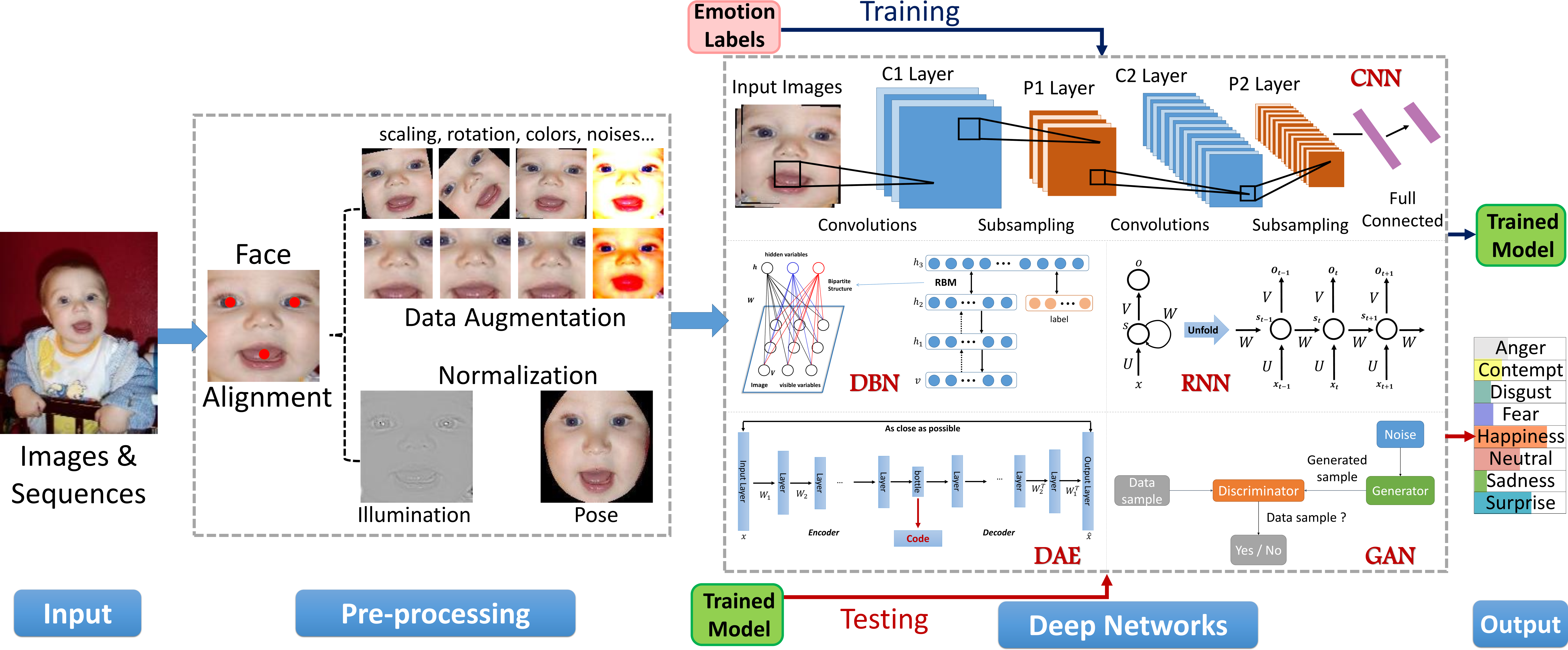}
		\caption{The general pipeline of deep facial expression recognition systems.}
		\label{system}
	\end{figure*}
	
	\subsection{Pre-processing}
	Variations that are irrelevant to facial expressions, such as different backgrounds, illuminations and head poses, are fairly common in unconstrained scenarios. Therefore, before training the deep neural network to learn meaningful features, pre-processing is required to align and normalize the visual semantic information conveyed by the face.
	
	\subsubsection{Face alignment}
	\begin{table}[t]
		\centering
		\footnotesize
		\renewcommand{\arraystretch}{1.1}
		\setlength{\tabcolsep}{1pt}
		\caption{Summary of different types of face alignment detectors that are widely used in deep FER models.}
		\label{alignment}
		\begin{tabular}{@{}|c|l|c|c|c|c|c|@{}}
			\hline
			\multicolumn{2}{|c|}{type} &\# points&real-time&speed &performance&used in \\\hline
			Holistic&AAM \cite{cootes2001active}  &68&\ding{55}  &fair  &\begin{tabular}[c]{@{}c@{}}poor \\generalization \end{tabular} &\cite{zeng2018facial2,hasani2017spatio}\\\hline
			\multirow{2}{*}{Part-based}&MoT \cite{zhu2012face}        &39/68&\ding{55}&\multirow{2}{*}{\begin{tabular}[c]{@{}c@{}}slow/\\fast\end{tabular}}  &\multirow{2}{*}{good}& \cite{kahou2013combining,devries2014multi}\\\cline{2-4}\cline{7-7}
			&DRMF \cite{asthana2013robust}       &66&\ding{55}&& &\cite{shin2016baseline,meng2017identity}\\\hline
			\multirow{3}{*}{\begin{tabular}[c]{@{}c@{}}Cascaded \\regression\end{tabular}}&SDM \cite{xiong2013supervised} &49  &\ding{51} &\multirow{3}{*}{\begin{tabular}[c]{@{}c@{}}fast/\\very fast\end{tabular}} &\multirow{3}{*}{\begin{tabular}[c]{@{}c@{}}good/\\very good\end{tabular}} &\cite{ng2015deep,jung2015joint}  \\\cline{2-4}\cline{7-7}
			&3000 fps \cite{ren2014face} &68&\ding{51}& &   &  \cite{hasani2017spatio}\\\cline{2-4}\cline{7-7}
			&Incremental \cite{asthana2014incremental} &49&\ding{51}&& &\cite{kim2017multi}\\\hline
			\multirow{2}{*}{\begin{tabular}[c]{@{}c@{}}Deep \\learning\end{tabular}}&cascaded CNN \cite{sun2013deep} &5&\ding{51} &\multirow{2}{*}{fast}  &\multirow{2}{*}{\begin{tabular}[c]{@{}c@{}}good/\\very good\end{tabular}} &\cite{zhang2017facial}\\\cline{2-4}\cline{7-7}
			&MTCNN \cite{zhang2016joint}     &5&\ding{51} & &   & \cite{yu2017deeper,yu2018spatio}\\\hline
		\end{tabular}
	\end{table}
	Face alignment is a traditional pre-processing step in many face-related recognition tasks. We list some well-known approaches and publicly available implementations that are widely used in deep FER. 
	
	Given a series of training data, the first step is to detect the face and then to remove background and non-face areas. The Viola-Jones (V\&J) face detector \cite{viola2001rapid} is a classic and widely employed implementation for face detection, which is robust and computationally simple for detecting near-frontal faces. 
	
	Although face detection is the only indispensable procedure to enable feature learning, further face alignment using the coordinates of localized landmarks can substantially enhance the FER performance \cite{mollahosseini2016going}. 
	This step is crucial because it can reduce the variation in face scale and in-plane rotation.
	Table \ref{alignment} investigates  facial landmark detection algorithms widely-used in deep FER and compares them in terms of efficiency and performance. The Active Appearance Model (AAM) \cite{cootes2001active} is a classic generative model that optimizes the required parameters from  holistic facial appearance and global shape patterns.  In discriminative models, 
	the mixtures of trees (MoT) structured models \cite{zhu2012face} and the discriminative response map fitting (DRMF) \cite{asthana2013robust} use part-based approaches that represent the face via the local appearance information around each landmark. Furthermore, a number of discriminative models directly use a cascade of regression functions to map the image appearance to landmark locations and have shown better results, e.g., the supervised descent method (SDM) \cite{xiong2013supervised} implemented in IntraFace \cite{ChuDC13}, the face alignment 3000 fps \cite{ren2014face}, and the incremental face alignment \cite{asthana2014incremental}. Recently, deep networks have been widely exploited for face alignment. Cascaded CNN \cite{sun2013deep} is the early work which predicts landmarks in a cascaded way. Based on this, Tasks-Constrained Deep Convolutional Network (TCDCN) \cite{zhang2014facial} and Multi-task CNN (MTCNN) \cite{zhang2016joint} further leverage multi-task learning to improve the performance.
	In general, cascaded regression has become the most popular and state-of-the-art methods for face alignment as its high speed and accuracy.

	In contrast to using only one detector for face alignment, some methods proposed to combine multiple detectors for better landmark estimation when processing faces in challenging unconstrained environments.
	Yu et al. \cite{yu2015image} concatenated three different facial landmark detectors to complement each other.
	Kim et al. \cite{kim2015hierarchical} considered different inputs (original image and histogram equalized image) and different face  detection  models (V\&J \cite{viola2001rapid} and MoT \cite{zhu2012face}), and the landmark set with the highest confidence provided by the Intraface \cite{ChuDC13} was selected.
	
	\subsubsection{Data augmentation}
	Deep neural networks require sufficient training data to ensure generalizability to a given recognition task. However, most publicly available databases for FER do not have a sufficient quantity of images for training. Therefore, data augmentation is a vital step for deep FER. Data augmentation techniques can be divided into two groups: on-the-fly data augmentation and offline data augmentation.
	
	Usually, the on-the-fly data augmentation is embedded in deep learning toolkits to alleviate overfitting. During the training step, the input samples are randomly cropped from the four corners and center of the image and then flipped horizontally, which can result in a dataset that is ten times larger than the original training data. Two common prediction modes are adopted during testing: only the center patch of the face is used for prediction (e.g., \cite{liu2017adaptive,meng2017identity}) or the prediction value is averaged over all ten crops (e.g., \cite{levi2015emotion,kim2015hierarchical}).
	
	Besides the elementary on-the-fly data augmentation, various offline  data augmentation operations have been designed to further expand data on both size and diversity. The most frequently used operations include random perturbations and transforms, e.g., rotation, shifting, skew, scaling, noise, contrast and color jittering. For example, common noise models, salt \& pepper and speckle noise \cite{pitaloka2017enhancing} and Gaussian noise \cite{lopes2017facial, zavarez2017cross} are employed to enlarge the data size. And for contrast transformation,  
	saturation and value (S and V components of the HSV color space) of each pixel are changed \cite{yu2017deeper} for data augmentation.
	Combinations of multiple operations can generate more unseen training samples and make the network more robust to deviated and rotated faces. 
	In \cite{li2015deep}, the authors  applied five image appearance filters (disk, average, Gaussian, unsharp and motion filters) and six affine transform matrices that were formalized by adding slight geometric transformations 
	to the identity matrix.  
	In \cite{yu2015image}, a more comprehensive  affine transform matrix was proposed to randomly generate images that varied in terms of rotation, skew and scale. 
	Furthermore, deep learning based technology can be applied for data augmentation. For example,
	a synthetic data generation system with 3D convolutional neural network (CNN) was created in \cite{abbasnejad2017using} to confidentially create faces with different levels of saturation in expression. 
	And the generative adversarial network (GAN) \cite{goodfellow2014generative} can also be applied to augment data by generating diverse appearances varying in poses and expressions. (see Section \ref{GAN}).

	\subsubsection{Face normalization}
	Variations in illumination and head poses can introduce large changes in images and hence impair the FER performance. Therefore, we introduce two typical face normalization methods to ameliorate these variations: illumination normalization and pose normalization (frontalization).
	\\\\
	\textbf{Illumination normalization:} Illumination and contrast can vary in different images even from the same person with the same expression, especially in unconstrained environments, which can result in large intra-class variances. 
	In \cite{shin2016baseline}, several frequently used illumination normalization algorithms, namely, isotropic diffusion (IS)-based normalization, discrete cosine transform (DCT)-based normalization \cite{chen2006illumination} and difference of Gaussian (DoG), were evaluated for illumination normalization.
	And \cite{li2015facial} employed homomorphic filtering based normalization, which has been reported to yield the most consistent results among all other techniques, to remove illumination normalization.
	Furthermore, related studies have shown that histogram equalization combined with illumination
	normalization results in better face recognition performance than that achieved using illumination normalization on it own.
	And many studies in the literature of deep FER (e.g., \cite{yu2015image,ebrahimi2015recurrent,bargal2016emotion,pitaloka2017enhancing}) have employed  histogram equalization to increase the global contrast of images for pre-processing. This method is effective when the brightness of the background and foreground are similar.
	However, directly applying histogram equalization may overemphasize local contrast.  To solve this problem, \cite{kuo2018compact} proposed a weighted summation approach to combine histogram equalization and linear mapping.
	And in \cite{pitaloka2017enhancing}, the authors compared three different methods: global contrast normalization (GCN), local normalization, and histogram equalization. GCN and histogram equalization were reported to achieve the best accuracy for the training and testing steps, respectively.
	\\\\
	\textbf{Pose normalization:}
	Considerable pose variation is another common and intractable problem in unconstrained settings. Some studies have employed pose normalization techniques to yield frontal facial views for FER (e.g., \cite{yao2016holonet,hu2017learning}), among which the most popular was proposed by Hassner et al. \cite{hassner2015effective}. Specifically, after localizing facial landmarks, a 3D texture reference model generic to all faces is generated to efficiently estimate visible facial components. Then, the initial frontalized face is synthesized by  back-projecting each input face image to the reference coordinate system.
	Alternatively, Sagonas et al. \cite{sagonas2015robust} proposed an effective statistical model to simultaneously localize landmarks and convert facial poses using only frontal faces.
	Very recently, a series of GAN-based deep models were proposed for frontal view synthesis (e.g., FF-GAN \cite{yin2017towards}, TP-GAN \cite{huang2017beyond}) and DR-GAN \cite{tran2017disentangled}) and report promising performances.
	
	\subsection{Deep networks for feature learning}
	Deep learning has recently become a hot research topic and has achieved state-of-the-art performance for a variety of applications \cite{deng2014deep}. Deep learning attempts to capture high-level abstractions through hierarchical architectures of multiple nonlinear transformations and representations. In this section, we briefly introduce some deep learning techniques that have been applied for FER. The traditional architectures of these deep neural networks are shown in Fig. \ref{system}.
	
	\subsubsection{Convolutional neural network (CNN)}
	CNN has been extensively used in diverse computer vision applications, including FER. At the beginning of the 21st century, several studies in the FER literature \cite{fasel2002robust,fasel2002head} found that the CNN is robust to face location changes and scale variations and behaves better than the multilayer perceptron (MLP) in the case of previously unseen face pose variations. \cite{matsugu2003subject} employed the CNN to address the problems of subject independence as well as translation, rotation, and scale invariance in the recognition of facial expressions.
	
	A CNN consists of three types of heterogeneous layers:  convolutional layers,  pooling layers, and fully connected layers. \textit{The convolutional layer} has a set of learnable filters to convolve through the whole input image and produce various specific types of activation feature maps. The convolution operation is associated with three main benefits: local connectivity, which learns correlations among neighboring pixels; weight sharing in the same feature map, which greatly reduces the number of the parameters to be learned; and shift-invariance to the location of the object. \textit{The pooling layer} follows the convolutional layer and is used to reduce the spatial size of the feature maps and the computational cost of the network. Average pooling and max pooling are the two most commonly used nonlinear down-sampling strategies for translation invariance. \textit{The fully connected layer} is usually included at the end of the network to ensure that all neurons in the layer are fully connected to activations in the previous layer and to enable the 2D feature maps to be converted into 1D feature maps for further feature representation and classification.
	
	
	We list the configurations and characteristics of some well-known CNN models that have been applied for FER in Table \ref{CNNs}. Besides these networks, several well-known derived frameworks also exist.
	In \cite{sun2015combining,sun2016facial}, region-based CNN (R-CNN) \cite{girshick2014rich} was used to learn features for FER. In \cite{li2017facial}, Faster R-CNN \cite{ren2015faster} was used to identify facial expressions by generating high-quality region proposals. Moreover, Ji et al. proposed 3D CNN \cite{ji20133d}  to capture motion information encoded in multiple adjacent frames for action recognition via 3D convolutions. Tran et al. \cite{tran2015learning} proposed the well-designed C3D, which exploits 3D convolutions on large-scale supervised training datasets to learn spatio-temporal features. Many related studies (e.g., \cite{fan2016video,nguyen2017deep}) have employed this network for FER involving image sequences.

	\begin{table}[t]
		\centering
		\caption{Comparison of CNN models and their achievements. DA = Data augmentation; BN = Batch normalization.}
		\label{CNNs}
		\begin{threeparttable}
			\begin{tabular}{@{}ccccc@{}}
				\toprule
				\multirow{2}{*}{} & AlexNet  & VGGNet  & GoogleNet & ResNet \\
				& \cite{krizhevsky2012imagenet}& \cite{simonyan2014very}&\cite{szegedy2015going}&\cite{he2016deep}\\\midrule
				Year & 2012 & 2014 & 2014 &2015  \\
				\# of layers\tnote{$\dagger$} & 5+3& 13/16 + 3 & 21+1 & 151+1 \\
				Kernel size\tnote{$\star$} & 11, 5, 3&3&7, 1, 3, 5&7, 1, 3, 5\\
				DA & \ding{51} & \ding{51}  & \ding{51}  & \ding{51}  \\
				Dropout& \ding{51}  & \ding{51} & \ding{51} & \ding{51}  \\
				Inception&  \ding{55} &  \ding{55} &  \ding{51} &\ding{55}   \\
				BN &  \ding{55} &  \ding{55} &   \ding{55} & \ding{51}  \\
				Used in & \cite{ouellet2014real} & \cite{levi2015emotion, ding2017facenet2expnet} & \cite{levi2015emotion, zhao2016peak} & \cite{hu2017learning, hasani2017facial}\\ \bottomrule
			\end{tabular}
			\begin{tablenotes}
				\item[$\dagger$] number of convolutional layers + fully connected layers
				\item[$\star$] size of the convolution kernel
			\end{tablenotes}
		\end{threeparttable}
	\end{table}
	
	\subsubsection{Deep belief network (DBN)} 
	DBN proposed by Hinton et al. \cite{hinton2006fast} is a graphical model that learns to extract a deep hierarchical representation of the training data. The traditional DBN is built with a stack of restricted Boltzmann machines (RBMs) \cite{hinton1986learning}, which are two-layer generative stochastic models composed of a visible-unit layer and a hidden-unit layer. These two layers in an RBM must form a bipartite graph
	without lateral connections. In a DBN, the units in higher layers are trained to learn the conditional dependencies among the units in the adjacent lower layers, except the top two layers, which have undirected connections. The training of a DBN contains two phases: pre-training and fine-tuning \cite{hinton2012practical}. First, an efficient layer-by-layer greedy learning strategy \cite{bengio2007greedy} is used to initialize the deep network in an unsupervised manner, which can prevent poor local optimal results to some extent without the requirement of a large amount of labeled data. During this procedure, contrastive divergence \cite{hinton2002training} is used to train RBMs in the DBN to estimate the approximation gradient of the log-likelihood. Then, the parameters of the network and the desired output are fine-tuned with a simple gradient descent under supervision.
	
	\subsubsection{Deep autoencoder (DAE)}
	DAE was first introduced in \cite{hinton2006reducing} to learn efficient codings for dimensionality reduction. In contrast to the previously mentioned networks, which are trained to predict target values, the DAE is optimized to reconstruct its inputs by minimizing the reconstruction error. Variations of the DAE exist, such as the denoising autoencoder \cite{vincent2010stacked}, which recovers the original undistorted input from partially corrupted data; the sparse autoencoder network (DSAE) \cite{le2013building}, which enforces sparsity on the learned feature representation; the contractive autoencoder (CAE$^1$) \cite{rifai2011contractive}, which adds an activity dependent regularization to induce locally invariant features; the convolutional autoencoder (CAE$^2$) \cite{masci2011stacked}, which uses convolutional (and
	optionally pooling) layers for the hidden layers in the network; and the variational auto-encoder (VAE) \cite{kingma2013auto}, which is a directed graphical model with certain types of latent variables to design complex generative models of data.
	
	\subsubsection{Recurrent neural network (RNN)}
	RNN is a connectionist model that captures temporal information and is more suitable for sequential data prediction with arbitrary lengths. In addition to training the deep neural network in a single feed-forward manner, RNNs include recurrent edges that span adjacent time steps and share the same parameters across all steps. The classic back propagation through time (BPTT) \cite{werbos1990backpropagation} is used to train the RNN.
	Long-short term memory (LSTM), introduced by Hochreiter \& Schmidhuber \cite{hochreiter1997long}, is a special form of the traditional RNN that is used to address the gradient vanishing and exploding problems that are common in training RNNs. The cell state in LSTM is regulated and controlled by three gates: an input gate that allows or blocks alteration of the cell state by the input signal, an output gate that enables or prevents the cell state to affect other neurons, and a forget gate that modulates the cell's self-recurrent connection to accumulate or forget its previous state. By combining these three gates, LSTM can model long-term dependencies in a sequence and has been widely employed for video-based expression recognition tasks.
	
	\subsubsection{Generative Adversarial Network (GAN)}
	
	GAN  was first introduced by Goodfellow et al  \cite{goodfellow2014generative} in 2014, which trains models through a minimax two-player game between a generator $\text{G}(\bf{z})$ that generates synthesized input data by mapping  latents $\bf{z}$ to data space with $\bf{z} \sim p(\bf{z})$ and a discriminator $\text{D}(\bf{x})$ that assigns probability $y=\text{Dis}(\bf{x}) \in [0,1]$ that $\bf{x}$ is an actual training sample to tell apart real from fake input data. The generator and the discriminator are trained alternatively and can both improve themselves by minimizing/maximizing the binary cross entropy $L_{GAN}=\log(\text{D}(\bf{x}))+\log(1-\text{D}(\text{G}(\bf{z})))$ with respect to D / G with $\bf{x}$ being a training sample and $\bf{z} \sim p(\bf{z})$. Extensions of GAN exist, such as the cGAN \cite{mirza2014conditional} that  adds a conditional information to control the output of the generator,  the DCGAN \cite{radford2015unsupervised} that adopts deconvolutional and convolutional neural networks to implement G and D respectively, the VAE/GAN \cite{larsen2015autoencoding} that uses learned feature representations in the GAN discriminator as basis for the VAE
	reconstruction objective, and the InfoGAN \cite{chen2016infogan} that can learn disentangled representations in a completely unsupervised manner.
	
	\subsection{Facial expression classification}
	After learning the deep features, the final step of FER is to classify the given face into one of the basic emotion categories.
	
	Unlike the traditional methods, where the feature extraction step and the feature classification step are independent, deep networks can perform FER in an end-to-end way. Specifically, a loss layer is added to the end of the network to regulate the back-propagation error; then, the prediction probability of each sample can be directly output by the network. In CNN, softmax loss is the most common used function that minimizes the cross-entropy between the estimated class probabilities and the ground-truth distribution. Alternatively, \cite{tang2013deep} demonstrated  the benefit of using a linear support vector machine (SVM) for the end-to-end training which  minimizes a margin-based loss instead of the cross-entropy. Likewise, \cite{dapogny2018investigating} investigated the adaptation of deep neural forests (NFs) \cite{kontschieder2015deep} which replaced the softmax loss layer with NFs and achieved competitive results for FER. 
	
	Besides the end-to-end learning way, another alternative is to employ the deep neural network (particularly a CNN) as a feature extraction tool and then apply additional independent classifiers, such as support vector machine  or random forest, to the extracted  representations \cite{donahue2014decaf,razavian2014cnn}. Furthermore, \cite{Otberdout2018DeepCD,acharya2018covariance} showed that the covariance descriptors computed on DCNN features and classification  with Gaussian kernels on Symmetric Positive Definite (SPD) manifold are more
	efficient than the standard classification with the softmax layer.

	\section{The state of the art }
	\label{method}
	In this section, we review the existing novel deep neural networks designed for FER and the related training strategies proposed to address expression-specific problems. We divide the works presented in the literature into two main groups depending on the type of data: deep FER networks for static images and deep FER networks for dynamic image sequences. 
	We then provide an overview of the current deep FER systems with respect to the network architecture and performance. Because some of the evaluated datasets do not provide explicit data groups for training, validation and testing, and the relevant studies may conduct experiments under different experimental conditions with different data, we summarize the expression recognition performance along with information about the data selection and grouping methods.
	
	\begin{table*}[htp]
		\scriptsize
		\renewcommand{\arraystretch}{1}
		\setlength{\tabcolsep}{1.1pt}
		\centering
		\caption{Performance summary of representative methods for static-based deep facial expression recognition on the most widely evaluated datasets.  Network size = depth \& number of parameters; Pre-processing = Face Detection \& Data Augmentation \& Face Normalization; IN = Illumination Normalization; $\mathcal{NE}$ = Network Ensemble; $\mathcal{CN}$ = Cascaded Network; $\mathcal{MN}$ = Multitask Network; LOSO = leave-one-subject-out.}
		\label{result1}
		\begin{threeparttable}
			\begin{tabular}{@{}
					|m{0.05\textwidth}<{\centering}
					|m{0.1\textwidth}<{\centering}
					|m{0.08\textwidth}<{\centering}|m{0.06\textwidth}<{\centering}
					|c|c
					|m{0.06\textwidth}<{\centering}|c|c
					|m{0.14\textwidth}<{\centering}
					|m{0.05\textwidth}<{\centering}
					|c
					|c|@{}}
				\hline
				Datasets
				& Method
				&\multicolumn{2}{c|}{\begin{tabular}[c]{@{}c@{}} Network\\ type\end{tabular}}
				&\multicolumn{2}{c|}{\begin{tabular}[c]{@{}c@{}} Network\\ size\end{tabular}}
				&\multicolumn{3}{c|}{Pre-processing}
				&Data selection 
				& Data group 
				&\begin{tabular}[c]{@{}c@{}}Additional \\classifier\end{tabular}
				& Performance\tnote{1}  (\%) \\ \hline\hline
				
				\multirow{15}{*}{\textbf{CK+}}
				
				&Ouellet  14 \cite{ouellet2014real} 
				&\multicolumn{2}{c|}{\begin{tabular}[c]{@{}c@{}} CNN (AlexNet)\end{tabular}}
				&-&-
				&V\&J&-&-
				&\multirow{2}{*}{the last frame} 
				&\multirow{2}{*}{LOSO} 
				&SVM
				&7 classes\tnote{$\dagger$} : (94.4)\\\cline{2-9}\cline{12-13}
				
				&Li et al. 15 \cite{li2015facial} 
				&\multicolumn{2}{c|}{RBM}
				&4&-
				&V\&J&-&IN
				&&
				&\ding{55}
				&
				6 classes: 96.8 \\\cline{2-13}
				
				&Liu et al. 14  \cite{liu2014facial}
				& DBN &$\mathcal{CN}$
				&6&2m
				&\ding{51}&-&-
				&\multirow{7}{*}{\begin{tabular}[c]{@{}c@{}}the last three frames\\  and the first frame\end{tabular}}
				&8 folds
				&AdaBoost
				&6 classes: 96.7\\\cline{2-9}\cline{11-13}
				
				& Liu et al. 13 \cite{liu2013aware}
				&CNN, RBM &$\mathcal{CN}$
				&5&-
				&V\&J&-&-
				& 
				&10 folds
				&SVM
				&8 classes: 92.05 (87.67)\\\cline{2-9}\cline{11-13}
				
				&Liu et al. 15 \cite{liu2015inspired} 
				&CNN, RBM  &$\mathcal{CN}$
				&5&-
				&V\&J&-&-
				&
				&10 folds
				&SVM
				&7 classes\tnote{$\ddagger$} : 93.70\\\cline{2-9}\cline{11-13}
				
				& \begin{tabular}[c]{@{}c@{}}Khorrami\\ et al. 15\end{tabular} \cite{khorrami2015deep} 
				&\multicolumn{2}{c|}{zero-bias CNN}
				&4&7m
				&\ding{51}&\ding{51}&-
				&  
				&10 folds
				&\ding{55}
				&6 classes: 95.7; 8 classes: 95.1\\\cline{2-9}\cline{11-13}
				
				& Ding et al. 17 \cite{ding2017facenet2expnet} 
				& CNN&fine-tune 
				&8&11m
				&IntraFace&\ding{51}&-
				&   
				&10 folds
				&\ding{55}
				&6 classes: (98.6); 8 classes: (96.8)\\\cline{2-13}

				& Zeng et al. 18 \cite{zeng2018facial2}
				&\multicolumn{2}{c|}{DAE (DSAE)}
				&3&-
				&AAM&-&-
				& \begin{tabular}[c]{@{}c@{}}the last four frames \\and the first frame\end{tabular} &LOSO
				&\ding{55}
				&\begin{tabular}[c]{@{}c@{}}7 classes\tnote{$\dagger$} : 95.79 (93.78) \\8 classes: 89.84 (86.82)\end{tabular}\\\cline{2-13}

				& Cai et al. 17 \cite{cai2017island}
				&CNN&loss layer
				&6&-
				&DRMF&\ding{51}&IN
				&\multirow{8}{*}{the last three frames}
				&10 folds
				&\ding{55} 
				&7 classes\tnote{$\dagger$} : 94.39 (90.66)\\ \cline{2-9}\cline{11-13}
				
				& Meng et al. 17 \cite{meng2017identity}
				&CNN&$\mathcal{MN}$
				&6&-
				&DRMF&\ding{51}&-
				&
				&8 folds
				&\ding{55}
				&7 classes\tnote{$\dagger$} : 95.37 (95.51)\\\cline{2-9}\cline{11-13}
				
				& Liu et al. 17 \cite{liu2017adaptive}
				& CNN&
				loss layer
				&11&-
				&IntraFace&\ding{51}&IN
				&  
				&8 folds
				&\ding{55} 
				&7 classes\tnote{$\dagger$} : 97.1 (96.1)\\\cline{2-9}\cline{11-13}
				
				& Yang et al. 18 \cite{yang2018facial}
				&\multicolumn{2}{c|}{GAN (cGAN)}
				&-&-
				&MoT&\ding{51}&-
				&
				&10 folds
				&\ding{55}
				&7 classes\tnote{$\dagger$} : 97.30 (96.57)\\\cline{2-9}\cline{11-13}
				
				&Zhang et al. 18 \cite{zhang2018From}
				& CNN&$\mathcal{MN}$
				&-&-
				&\ding{51}&\ding{51}&-
				&
				&10 folds
				&\ding{55}
				&6 classes: 98.9\\\hline\hline
				
				\multirow{4}{*}{\textbf{JAFFE}}
				
				&Liu et al. 14 \cite{liu2014facial}
				&DBN&$\mathcal{CN}$
				&6&2m
				&\ding{51}&-&-
				&\multirow{4}{*}{213 images} 
				&\multirow{2}{*}{LOSO }
				&AdaBoost
				&7 classes\tnote{$\ddagger$} : 91.8\\\cline{2-9}\cline{12-13}
				
				& \begin{tabular}[c]{@{}c@{}}Hamester \\et al. 15\end{tabular} \cite{hamester2015face}
				&CNN, CAE& $\mathcal{NE}$ 
				&3&-
				&-&-&IN
				&
				&
				&\ding{55}
				&7 classes\tnote{$\ddagger$} : (95.8)\\\hline\hline
				
				\multirow{8}{*}{\textbf{MMI}}
				
				&Liu et al. 13 \cite{liu2013aware}
				&CNN, RBM&$\mathcal{CN}$
				&5&-
				&V\&J&-&-
				&\multirow{2}{*}{\begin{tabular}[c]{@{}c@{}}the middle three frames\\  and the first frame\end{tabular}} 
				&10 folds
				&SVM
				&7 classes\tnote{$\ddagger$} : 74.76 (71.73)\\\cline{2-9}\cline{11-13}
				
				&Liu et al. 15 \cite{liu2015inspired} 
				&CNN, RBM &$\mathcal{CN}$
				&5&-
				&V\&J&-&-
				&
				&10 folds
				&SVM
				&7 classes\tnote{$\ddagger$} : 75.85\\\cline{2-13}
				
				&\begin{tabular}[c]{@{}c@{}}Mollahosseini \\et al. 16\end{tabular} \cite{mollahosseini2016going}
				&\multicolumn{2}{c|}{CNN (Inception)}
				&11&7.3m
				&IntraFace&\ding{51}&-
				&images from each sequence  
				& 5 folds  
				&\ding{55}
				&6 classes: 77.9\\ \cline{2-13}
				
				&Liu et al. 17 \cite{liu2017adaptive}
				& CNN& 
				loss layer
				&11&-
				&IntraFace&\ding{51}&IN
				&\multirow{4}{*}{the middle three  frames}
				&10 folds
				&\ding{55}
				& 6 classes: 78.53 (73.50) \\\cline{2-9}\cline{11-13}
				& Li et al. 17 \cite{li2017reliable}
				&CNN&loss layer   
				& 8&5.8m
				&IntraFace&\ding{51}&-
				&
				&5 folds
				&SVM
				&6 classes: 78.46\\\cline{2-9}\cline{11-13}
				
				&Yang et al. 18 \cite{yang2018facial}
				&\multicolumn{2}{c|}{GAN (cGAN)}
				&-&-
				&MoT&\ding{51}&-
				&
				&10 folds
				&\ding{55}
				&6 classes: 73.23 (72.67)\\\hline\hline

				\multirow{6}{*}{\textbf{TFD}}
				
				
				&  Reed et al. 14 \cite{reed2014learning}
				&RBM& $\mathcal{MN}$ 
				&-&-
				&-&-&-
				&\begin{tabular}[c]{@{}c@{}}4,178 emotion labeled\\  3,874 identity labeled\end{tabular}
				&\multirow{6}{*}{\begin{tabular}[c]{@{}c@{}}5 official \\folds\end{tabular}}
				&SVM
				&Test: 85.43\\\cline{2-10}\cline{12-13}
				
				&Devries et al. 14 \cite{devries2014multi} 
				& CNN& $\mathcal{MN}$
				&4&12.0m
				&MoT&\ding{51}&IN
				&\multirow{6}{*}{4,178 labeled images} 
				&
				&\ding{55}
				&\begin{tabular}[c]{@{}c@{}}Validation: 87.80 \\Test: 85.13 (48.29)\end{tabular}\\\cline{2-9}\cline{12-13}
				
				
				&\begin{tabular}[c]{@{}c@{}}Khorrami\\ et al. 15 \end{tabular}\cite{khorrami2015deep} 
				&\multicolumn{2}{c|}{zero-bias CNN}
				&4& 7m 
				&\ding{51}&\ding{51}&-
				&
				& 
				&\ding{55}
				&Test: 88.6\\\cline{2-9}\cline{12-13}
				
				&Ding et al. 17 \cite{ding2017facenet2expnet}
				&CNN& fine-tune 
				&8&11m
				&IntraFace&\ding{51}&-
				&
				&
				&\ding{55}
				&Test: 88.9 (87.7)\\\hline\hline

				\multirow{7}{*}{\textbf{\begin{tabular}[c]{@{}c@{}}FER \\2013\end{tabular}}}
				
				& Tang 13 \cite{tang2013deep} 
				&CNN& loss layer
				&4&12.0m
				&-&\ding{51}&IN
				& \multicolumn{2}{c|}{\multirow{10}{*}{\begin{tabular}[c]{@{}c@{}}Training Set: 28,709\\Validation Set: 3,589 \\Test Set: 3,589 \end{tabular}}} 
				&\ding{55}
				&Test: 71.2\\\cline{2-9}\cline{12-13}
				
				&Devries et al. 14 \cite{devries2014multi} 
				& CNN&$\mathcal{MN}$
				&4&12.0m
				&MoT&\ding{51}&IN
				&\multicolumn{2}{c|}{} 
				&\ding{55}
				&Validation+Test: 67.21\\\cline{2-9}\cline{12-13}
				
				&Zhang et al. 15 \cite{Zhang2015Learning}
				&CNN&$\mathcal{MN}$
				&6&21.3m
				&SDM&-&-
				&\multicolumn{2}{c|}{}    
				&\ding{55}
				&Test: 75.10\\\cline{2-9}\cline{12-13}
				
				&Guo et al. 16 \cite{guo2016deep}
				&CNN&loss layer
				&10&2.6m
				&SDM&\ding{51}&-
				& \multicolumn{2}{c|}{} 
				&k-NN
				&Test: 71.33\\\cline{2-9}\cline{12-13}
				
				&Kim et al. 16 \cite{kim2016fusing} 
				&CNN&$\mathcal{NE}$ 
				&5&2.4m
				&IntraFace&\ding{51}&IN
				& \multicolumn{2}{c|}{}    
				&\ding{55}
				&Test: 73.73\\\cline{2-9}\cline{12-13}
				
				& \begin{tabular}[c]{@{}c@{}}pramerdorfer \\et al. 16 \end{tabular}\cite{pramerdorfer2016facial} 
				&CNN&$\mathcal{NE}$ 
				&10/16/33&\begin{tabular}[c]{@{}c@{}}1.8/1.2/5.3\\(m)
				\end{tabular}
				&-&\ding{51}&IN
				& \multicolumn{2}{c|}{}
				&\ding{55}
				&Test:75.2\\\hline\hline
				

				\multirow{15}{*}{\textbf{\begin{tabular}[c]{@{}c@{}}SFEW \\2.0\end{tabular}}}
				
				&levi et al. 15  \cite{levi2015emotion}
				&CNN&$\mathcal{NE}$ 
				&\multicolumn{2}{c|}{\begin{tabular}[c]{@{}c@{}}VGG-S/VGG-M/\\GoogleNet\end{tabular}}
				&MoT&\ding{51}&-
				&\multicolumn{2}{c|}{\begin{tabular}[c]{@{}c@{}}891 training, 431 validation, \\and 372 test\end{tabular}} 
				&\ding{55}
				&\begin{tabular}[c]{@{}c@{}}Validation: 51.75\\Test: 54.56 \end{tabular}\\\cline{2-13}
				
				&Ng et al. 15 \cite{ng2015deep}
				&CNN& fine-tune
				&\multicolumn{2}{c|}{AlexNet}
				&IntraFace&\ding{51}&-
				&\multicolumn{2}{c|}{\begin{tabular}[c]{@{}c@{}}921 training, ? validation, \\and 372 test\end{tabular}} 
				&\ding{55}
				&\begin{tabular}[c]{@{}c@{}}Validation: 48.5 (39.63)\\Test: 55.6 (42.69) \end{tabular}\\\cline{2-13}
				
				&Li et al. 17 \cite{li2017reliable}
				&CNN& loss layer
				&8&5.8m
				&IntraFace&\ding{51}&-
				& \multicolumn{2}{c|}{921 training, 427 validation} 
				&SVM
				&Validation: 51.05\\\cline{2-13}
				
				& Ding et al. 17 \cite{ding2017facenet2expnet}
				&CNN&fine-tune
				&8&11m
				&IntraFace&\ding{51}&-
				&\multicolumn{2}{c|}{891 training, 425 validation}
				&\ding{55}
				&Validation: 55.15 (46.6)\\\cline{2-13}

				&Liu et al. 17 \cite{liu2017adaptive} 
				&CNN& 
				loss layer
				&11&-
				&IntraFace&\ding{51}&IN
				& \multicolumn{2}{c|}{\multirow{9}{*}{\begin{tabular}[c]{@{}c@{}}958training, \\436 validation, \\and 372 test\end{tabular}}} 
				& \ding{55}
				&Validation: 54.19 (47.97)\\\cline{2-9}\cline{12-13}
				
				& Cai et al. 17 \cite{cai2017island}
				&CNN&loss layer
				&6&-
				& DRMF&\ding{51}&IN
				&\multicolumn{2}{c|}{} 
				&\ding{55}
				&\begin{tabular}[c]{@{}c@{}}Validation: 52.52 (43.41)\\Test: 59.41 (48.29)\end{tabular}\\\cline{2-9}\cline{12-13}
				
				&Meng et al. 17 \cite{meng2017identity}
				&CNN&$\mathcal{MN}$
				&6&-
				&DRMF&\ding{51}&-
				&\multicolumn{2}{c|}{} 
				&\ding{55}
				&\begin{tabular}[c]{@{}c@{}}Validation: 50.98 (42.57)\\Test: 54.30 (44.77)\end{tabular}\\\cline{2-9}\cline{12-13}
				
				&Kim et al. 15 \cite{kim2015hierarchical}
				&CNN&$\mathcal{NE}$
				&5&-
				&multiple&\ding{51}&IN
				&\multicolumn{2}{c|}{} 
				&\ding{55}&\begin{tabular}[c]{@{}c@{}}Validation: 53.9\\Test: 61.6\end{tabular}\\ \cline{2-9}\cline{12-13}
				
				&Yu et al. 15 \cite{yu2015image}
				&CNN&$\mathcal{NE}$
				&8&6.2m
				&multiple&\ding{51}&IN
				&\multicolumn{2}{c|}{} 
				&\ding{55}
				&\begin{tabular}[c]{@{}c@{}}Validation: 55.96 (47.31)\\Test: 61.29 (51.27)\end{tabular}\\ \hline
			\end{tabular}
			\begin{tablenotes}
				\item[1] The value in parentheses is the mean accuracy, which is calculated with the confusion matrix given by the authors.
				\item[$\dagger$] 7 Classes: Anger, Contempt, Disgust, Fear, Happiness, Sadness, and Surprise.
				\item[$\ddagger$] 7 Classes: Anger, Disgust, Fear, Happiness, Neutral, Sadness, and Surprise.
			\end{tablenotes}
		\end{threeparttable}
	\end{table*}
	
	\subsection{Deep FER networks for static images}
	\label{sta}
	A large volume of the existing studies conducted expression recognition  tasks based on static images without considering temporal information due to the convenience of data processing and the availability of the relevant training and test material. We first introduce specific pre-training and fine-tuning skills for FER, then review the novel deep neural networks in this field. 
	For each of the most frequently evaluated datasets, Table \ref{result1} shows the current state-of-the-art methods in the field that are explicitly conducted in a person-independent protocol (subjects in the training and testing sets are separated).
	
	\subsubsection{Pre-training and fine-tuning}
	
	As mentioned before, direct training of deep networks on relatively small facial expression datasets is prone to overfitting. To mitigate this problem, many studies used additional task-oriented data to pre-train their self-built networks from scratch or fine-tuned on well-known pre-trained models (e.g., AlexNet \cite{krizhevsky2012imagenet}, VGG\cite{simonyan2014very}, VGG-face \cite{parkhi2015deep} and GoogleNet \cite{szegedy2015going}). Kahou et al. \cite{kahou2013combining,kaneko2016adaptive} indicated that the use of additional data can help to obtain models with high capacity without overfitting, thereby enhancing the FER performance.
	
	To select appropriate auxiliary data,  large-scale face recognition (FR) datasets (e.g., CASIA WebFace \cite{yi2014learning}, Celebrity Face in the Wild (CFW) \cite{zhang2012finding}, FaceScrub dataset \cite{ng2014data}) or relatively large FER datasets (FER2013 \cite{goodfellow2013challenges} and TFD \cite{susskind2010toronto}) are suitable.
	Kaya et al. \cite{kaya2017video} suggested that VGG-Face which was trained for FR overwhelmed ImageNet which was developed for objected recognition.
	Another interesting result observed by Knyazev et al. \cite{knyazev2017convolutional} is that pre-training on larger FR  data positively affects the emotion recognition accuracy, and further fine-tuning with additional FER datasets can help improve the performance.
	
	\begin{figure}
		\includegraphics[width=8.5cm]{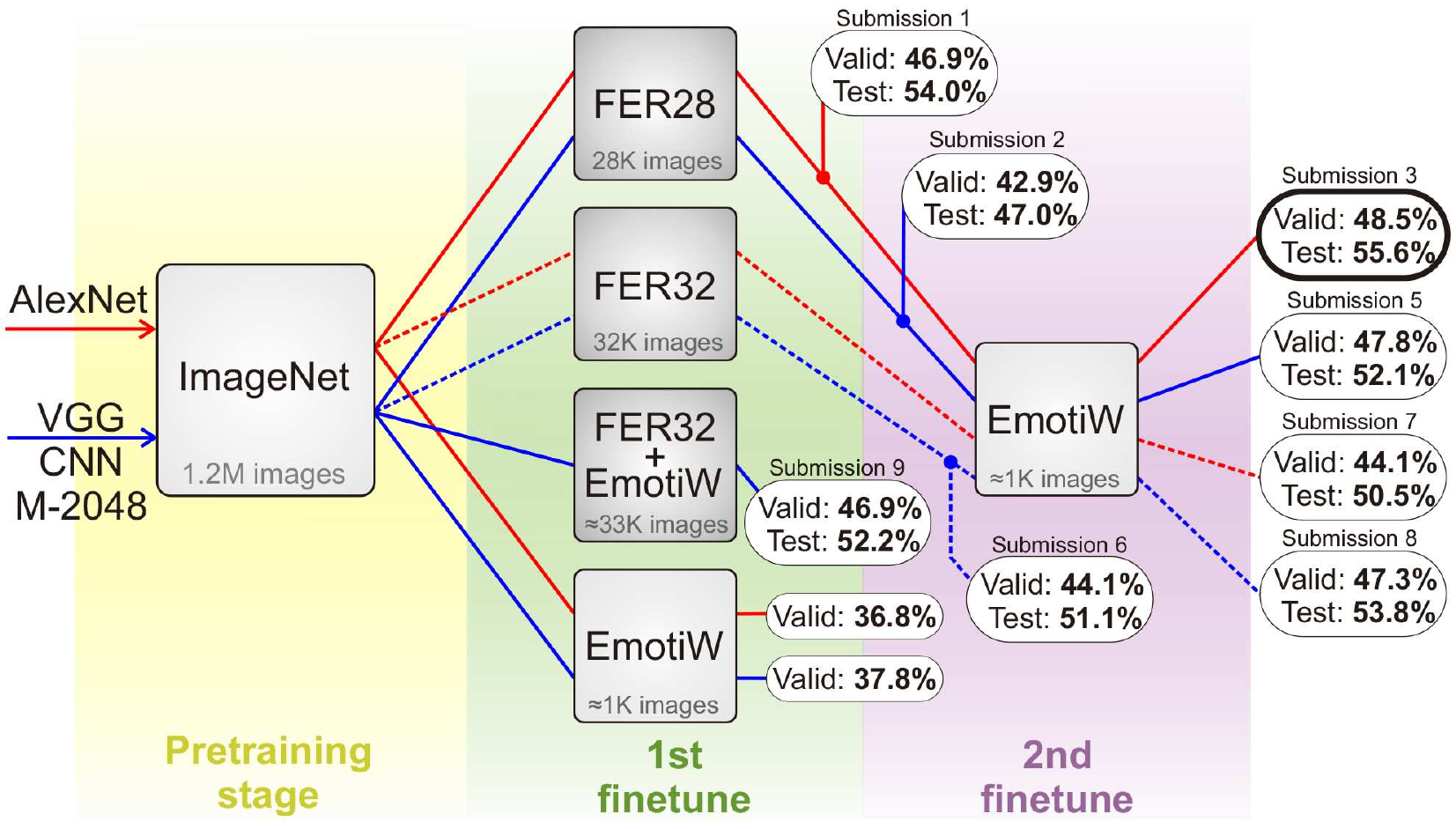}
		\caption{Flowchart of the different fine-tuning combinations used in \cite{ng2015deep}. Here, ``FER28'' and ``FER32'' indicate different parts of the FER2013 datasets. ``EmotiW'' is the target dataset.  The proposed two-stage fine-tuning strategy (Submission 3) exhibited the best performance.}
		\label{finetune}
	\end{figure}
	\begin{figure}
		\setlength{\abovecaptionskip}{1pt}
		\includegraphics[width=8.5cm]{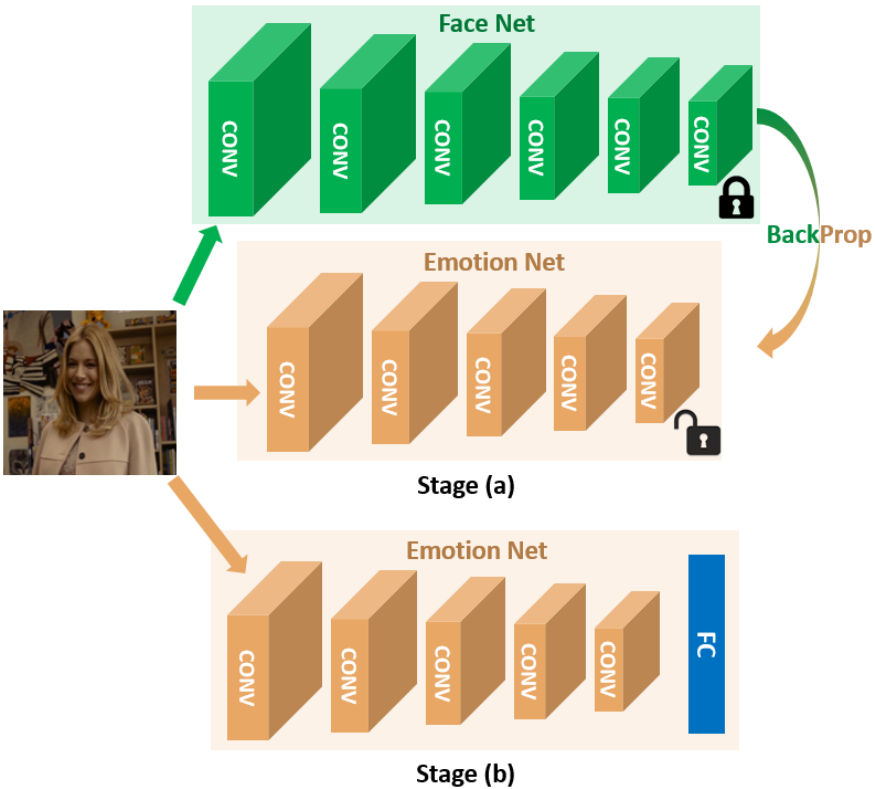}
		\caption{Two-stage training flowchart in \cite{ding2017facenet2expnet}. In stage (a), the deeper face net is frozen and provides the feature-level  regularization that pushes the convolutional  features of the expression net to be close to the face net by using the proposed distribution function. Then, in stage (b), to further improve the discriminativeness of the learned features, randomly initialized fully connected layers are added and jointly trained with the whole expression net using the expression label information. }
		\label{face2exp}
	\end{figure}
	Instead of directly using the pre-trained or fine-tuned models to extract features on the target dataset,  a multistage fine-tuning strategy \cite{ng2015deep} (see ``Submission 3'' in Fig. \ref{finetune}) can achieve better performance: after the first-stage fine-tuning using FER2013 on pre-trained models, a second-stage fine-tuning based on the training part of the target dataset (EmotiW) is employed to refine the models to adapt to a more specific dataset (i.e., the target dataset). 
	
	Although pre-training and fine-tuning on external FR data can indirectly avoid the problem of small training data, the networks are trained separately from the FER and the face-dominated information remains in the learned features which may weaken the network’s ability to represent expressions.
	To eliminate this effect, a two-stage training algorithm FaceNet2ExpNet \cite{ding2017facenet2expnet} was proposed (see Fig. \ref{face2exp}). The fine-tuned face net serves as a good initialization for the expression net and is used to guide the learning of the convolutional layers only. And the fully connected layers are trained from scratch with expression information to regularize the training of the target FER net.
	
	\subsubsection{Diverse network input}
	\label{input}
	Traditional practices commonly use the whole aligned face of RGB images as the input of the network to learn features for FER. 
	However, these raw data lack important information, such as homogeneous or regular textures and invariance in terms of image scaling, rotation, occlusion and illumination, which may represent confounding factors for FER. Some methods have employed diverse handcrafted features and their extensions as the network input to alleviate this problem.
	
	\begin{figure}[t]
		\setlength{\abovecaptionskip}{2pt}
		\includegraphics[width=8.5cm,height=3.5cm]{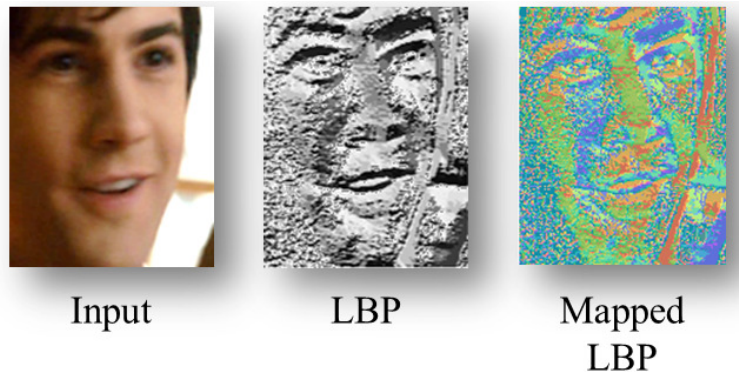}
		\caption{Image intensities (left) and LBP codes (middle). \cite{levi2015emotion} proposed mapping these values to a 3D metric space (right) as the input of CNNs. }
		\label{lbp}
	\end{figure}
	
	Low-level representations encode features from small regions in the given RGB image, then cluster and pool these features with local histograms, which are robust to illumination variations and small registration errors. A novel mapped LBP feature \cite{levi2015emotion} (see Fig. \ref{lbp}) was proposed  for illumination-invariant FER. Scale-invariant feature transform (SIFT) \cite{lowe1999object}) features that are robust against image scaling and rotation are employed \cite{zhang2016deep} for multi-view FER tasks.
	Combining different descriptors in outline, texture, angle, and color as the input data can also help enhance the deep network performance \cite{luo2017facial, zeng2018facial2}.
	
	Part-based representations extract features according to the target task, which remove noncritical parts from the whole image and exploit key parts that are sensitive to the task.
	\cite{chen2018softmax} indicated that three regions of interest (ROI), i.e., eyebrows, eyes and mouth, are strongly related to facial expression changes, and cropped these regions as the input of DSAE. 
	Other researches proposed to automatically learn the key parts for facial expression. For example,  \cite{mavani2017facial} employed a deep multi-layer network \cite{cornia2016deep} to detect the saliency map which put intensities on parts demanding visual attention.  And  \cite{wu2018adaptive} applied the neighbor-center difference vector (NCDV) \cite{lu2015cost} to obtain features with more intrinsic information.

	\subsubsection{Auxiliary  blocks \& layers}
	\begin{figure}[tp]
		\subfigure[Three different supervised blocks in \cite{hu2017learning}. SS\_Block for shallow-layer supervision, IS\_Block for intermediate-layer supervision, and DS\_Block for deep-layer supervision.]{\label{blocks}
			\includegraphics[width=8.5cm]{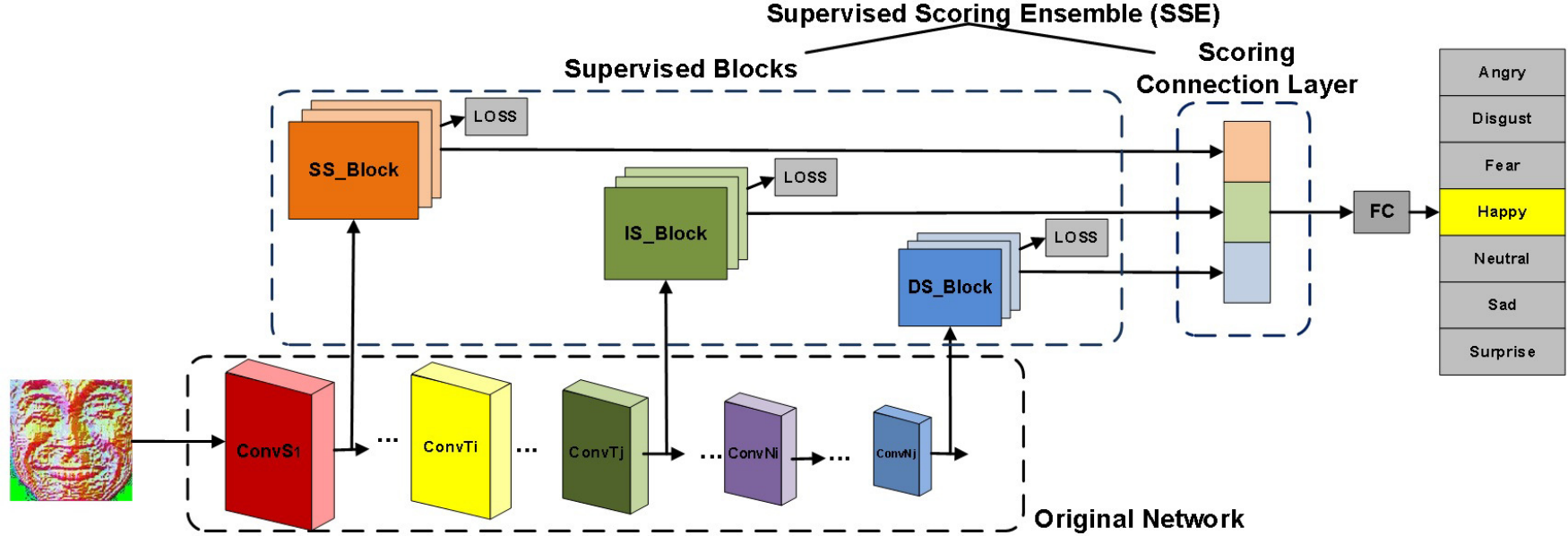}}
		\subfigure[Island loss layer in \cite{cai2017island}. The island loss calculated at the feature extraction layer and the softmax loss calculated at the decision layer are combined to supervise the CNN training.]{\label{island}
			\includegraphics[width=8.5cm]{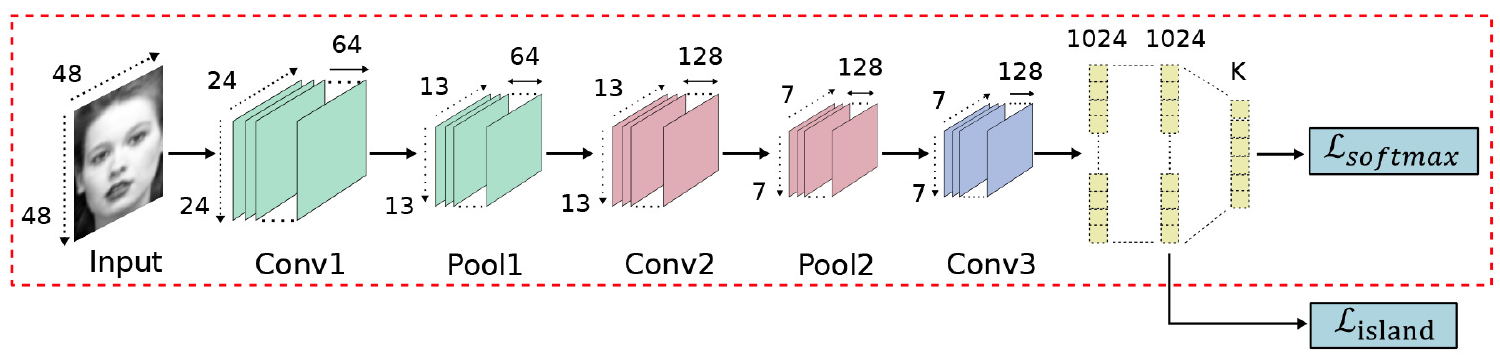}}
		\subfigure[(N+M)-tuple clusters loss layer in \cite{liu2017adaptive}. During training, the identity-aware hard-negative mining
		and online positive mining schemes are used to decrease the inter-identity variation in the same expression class.]{\label{loss}
			\includegraphics[width=8.5cm]{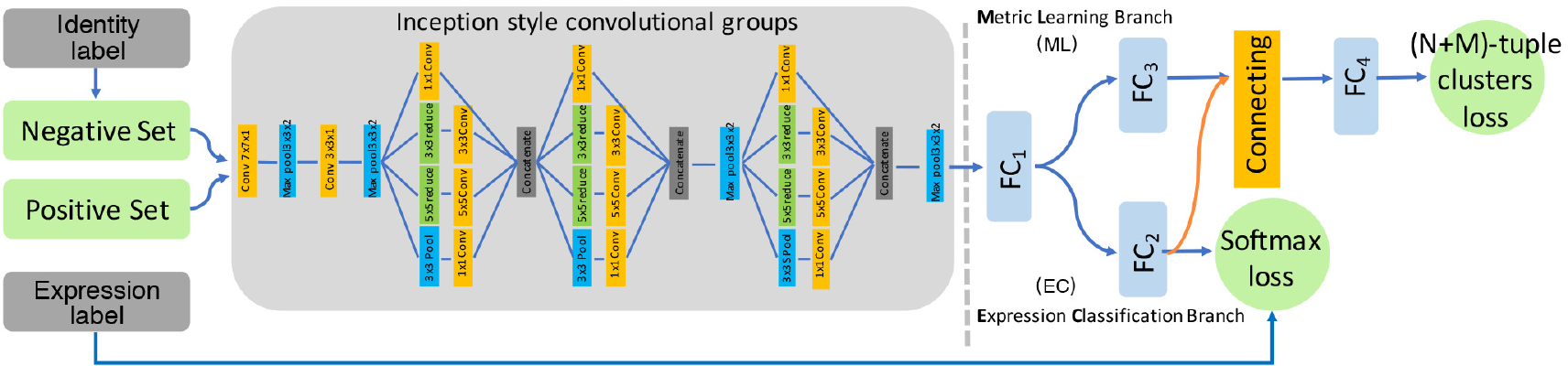}}
		\label{auxiliary}
		\caption{Representative functional layers or blocks that are specifically designed for deep facial expression recognition.}
	\end{figure}
	Based on the foundation architecture of CNN, several studies have proposed the addition of well-designed auxiliary blocks or layers to enhance the expression-related representation capability of learned features.
	
	Based on the foundation architecture of CNN, several studies have proposed the addition of well-designed auxiliary blocks or layers to enhance the expression-related representation capability of learned features.
	
	A novel CNN architecture, HoloNet \cite{yao2016holonet}, was designed for FER, where CReLU \cite{shang2016understanding} was combined with the powerful residual structure \cite{he2016deep} to increase the network depth without efficiency reduction and an inception-residual block \cite{szegedy2016rethinking,szegedy2017inception} was uniquely designed for FER to learn multi-scale features to capture variations in expressions. 
	Another CNN model, Supervised Scoring Ensemble (SSE) \cite{hu2017learning}, was introduced to enhance the supervision degree for FER, where three types of supervised blocks  were embedded in the early hidden layers of the mainstream CNN for shallow, intermediate and deep supervision, respectively (see Fig. \ref{blocks}).
	
	And a feature selection network (FSN) \cite{zhaofeature} was designed by embedding a feature selection mechanism inside the AlexNet, which  automatically filters irrelevant features and emphasizes correlated features according to learned feature maps of facial expression.
	Interestingly, Zeng et al. \cite{zeng2018facial} pointed out that the  inconsistent annotations among different FER databases are  inevitable which would damage the performance  when the training set is enlarged by merging multiple datasets. To address this problem, the authors proposed  an Inconsistent Pseudo Annotations to Latent Truth (IPA2LT) framework. In IPA2LT, an end-to-end trainable LTNet is designed to discover the latent truths from the human annotations and the machine annotations trained from different datasets  by maximizing the log-likelihood of these inconsistent annotations.
	
	The traditional softmax loss layer in CNNs simply forces features of different classes to remain apart, but FER  in real-world scenarios suffers from not only high inter-class similarity but also high intra-class variation. Therefore, several works have proposed novel loss layers for FER. Inspired by the center loss \cite{wen2016discriminative}, which penalizes the distance between deep features and their corresponding class centers, two variations  were  proposed to assist the supervision of the softmax loss for more discriminative features for FER:  (1) island loss \cite{cai2017island} was formalized to further increase the pairwise distances between different  class centers (see Fig. \ref{island}), and (2) locality-preserving loss (LP loss) \cite{li2017reliable} was formalized to pull the locally neighboring features of the same class together so that the intra-class local clusters of each class are compact. Besides, based on the triplet loss \cite{schroff2015facenet}, which requires one positive example to be closer to the anchor than one negative example with a fixed gap, two variations  were proposed to replace or assist the supervision of the softmax loss: (1) exponential triplet-based loss \cite{guo2016deep} was formalized to give difficult samples more weight when updating the network, and (2) (N+M)-tuples cluster loss \cite{liu2017adaptive} was formalized to alleviate the difficulty of anchor selection and threshold validation in the triplet loss for identity-invariant FER (see Fig. \ref{loss} for details).
	Besides, a feature loss \cite{zeng2018hand} was proposed to provide complementary information for the deep feature during early training stage.

	\subsubsection{Network ensemble}
	Previous research suggested that assemblies of multiple networks can outperform an individual network \cite{ciregan2012multi}. Two key factors should be considered when implementing network ensembles: (1) sufficient diversity of the networks to ensure complementarity, and (2) an appropriate ensemble method that can effectively aggregate the committee networks.
	
	In terms of the first factor, different kinds of training data  and various network parameters  or architectures are considered to generate diverse committees. 
	Several pre-processing methods \cite{kim2016fusing}, such as deformation and normalization, and methods described in Section \ref{input} can generate different data to train diverse networks.
	By changing the size of filters, the number of neurons and the number of layers of the networks, and applying multiple random seeds for weight initialization, the diversity of the networks can also be enhanced \cite{kim2015hierarchical,pons2017supervised}.
	Besides, different architectures of networks can be used to enhance the diversity. For example, a CNN trained in a supervised way and a convolutional autoencoder  (CAE) trained in an unsupervised way were combined for network ensemble \cite{hamester2015face}.
	
	\begin{table}[tp]
		\centering
		\caption{Three primary ensemble methods on the decision level.}
		\label{decision}
		\begin{tabular}{@{}|m{0.07\textwidth}<{\centering}|m{0.25\textwidth}<{\centering}|m{0.09\textwidth}<{\centering}|@{}}\hline
			& definition & used in (example) \\\hline
			Majority Voting &  determine the class with the most votes using the predicted label yielded from each individual  &\cite{kim2015hierarchical,kim2016hierarchical,kim2016fusing}  \\\hline
			Simple Average  &  determine the class with the highest mean score using the posterior class probabilities yielded from each individual with the same weight&\cite{kim2015hierarchical,kim2016hierarchical,kim2016fusing}
			\\\hline
			Weighted Average&determine the class with the highest weighted mean score using the posterior class probabilities yielded from each individual with different weights & \cite{kahou2013combining,levi2015emotion,pramerdorfer2016facial,kaya2017video}\\\hline
		\end{tabular}
	\end{table}
	
\begin{figure}[tp]
		\subfigure[Feature-level ensemble in \cite{bargal2016emotion}. Three different features (\textit{fc5} of VGG13 + \textit{fc7} of VGG16 + \textit{pool} of Resnet) after normalization are concatenated to create a single feature vector (FV) that describes the input frame.]{\label{ensemble1}
			\includegraphics[width=8.5cm]{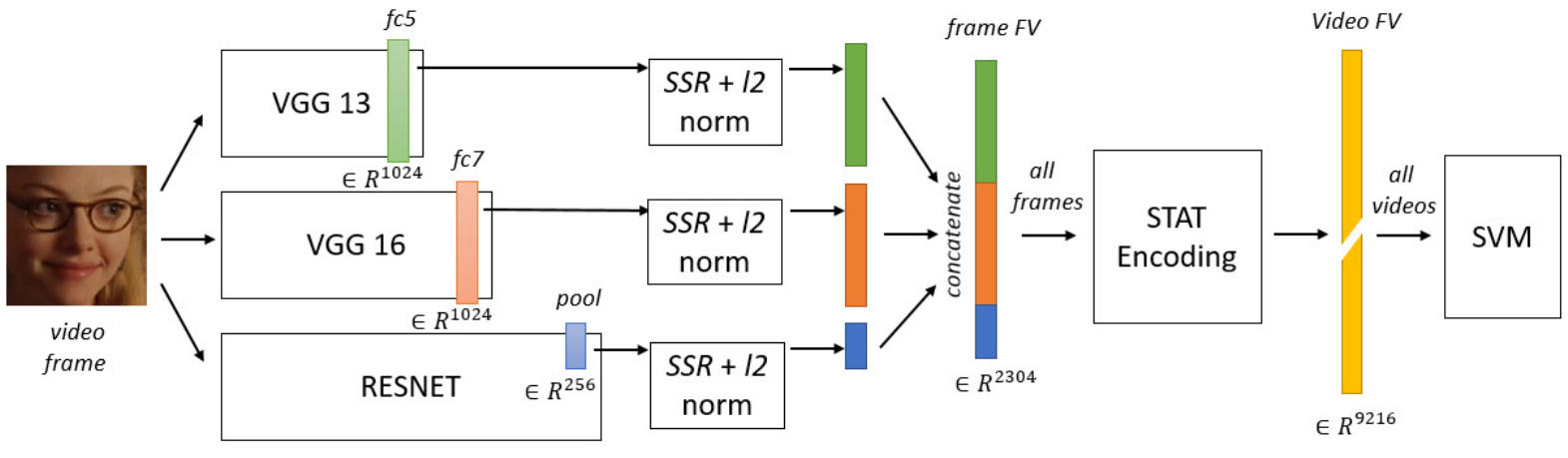}}
		\subfigure[Decision-level ensemble in \cite{kim2015hierarchical}. A 3-level hierarchical committee architecture with hybrid decision-level fusions was proposed to obtain sufficient decision diversity.]{\label{ensemble2}
			\includegraphics[width=8.5cm]{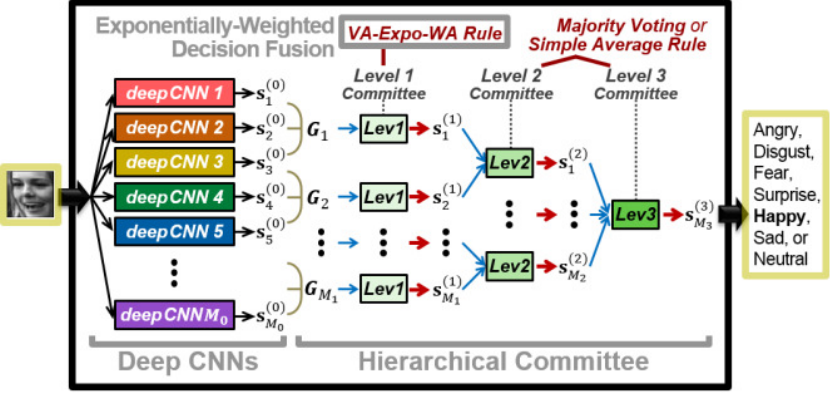}}
		\label{ensemble}
		\caption{Representative network ensemble systems at the feature level and decision level.}
	\end{figure}

	For the second factor, each member of the committee networks can be assembled at two different levels: the feature level and the decision level. For \textit{feature-level} ensembles, the most commonly adopted strategy is to concatenate features learned from different networks \cite{liu2016facial, bargal2016emotion}. 
	For example, \cite{bargal2016emotion} concatenated features learned from different networks  to obtain a single feature vector to describe the input image (see Fig. \ref{ensemble1}). For  \textit{decision-level} ensembles, three widely-used rules are applied: majority voting, simple average and weighted average. A summary of these three methods is provided in Table \ref{decision}. Because the weighted average rule considers the  importance and confidence of each individual, many weighted average methods have been proposed to find an optimal set of weights for network ensemble.
	\cite{kahou2013combining} proposed a random search method to  weight the model predictions for each emotion type. 
	\cite{yu2015image} used the log-likelihood loss and hinge loss to adaptively assign different weights to each network.
	\cite{kim2015hierarchical} proposed an exponentially weighted average based on the validation accuracy to emphasize qualified individuals (see Fig. \ref{ensemble2}). 
	\cite{pons2017supervised} used a CNN to learn weights for each individual model.
	
	\subsubsection{Multitask networks}
	\label{mscnn}
	Many existing networks for FER focus on a single task and learn features that are sensitive to expressions without considering interactions among other latent factors.  However, in the real world, FER is intertwined with various factors, such as head pose, illumination, and subject identity (facial morphology).  To solve this problem, multitask leaning is introduced to  transfer knowledge from other relevant tasks and to disentangle nuisance factors.
	
	Reed et al. \cite{reed2014learning} constructed a higher-order Boltzmann machine (disBM) to learn manifold coordinates for the relevant factors of expressions and proposed training strategies for disentangling so that the expression-related hidden units are invariant to face morphology.
	Other works \cite{devries2014multi, pons2018multi} suggested that simultaneously conducted FER with other tasks, such as facial landmark localization and  facial AUs \cite{ekman1997face} detection, can jointly improve FER performance.
	
	\begin{figure}
		\includegraphics[width=8.5cm]{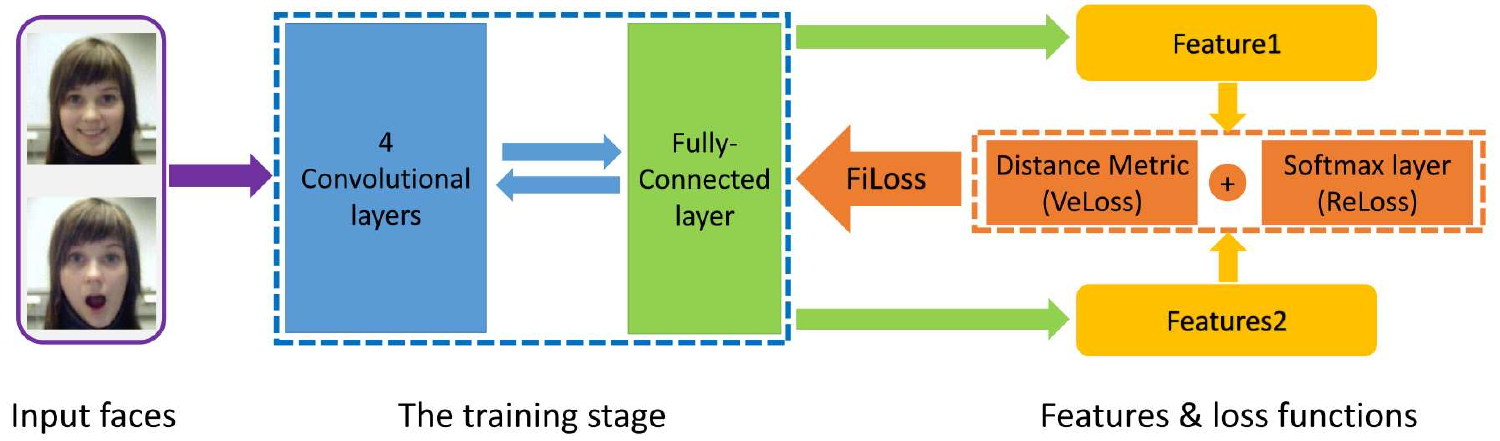}
		\caption{Representative multitask network for FER. In the proposed MSCNN \cite{zhang2017facial},  a pair of images is sent into the MSCNN during training. The expression recognition  task with cross-entropy loss, which learns features with large between-expression variation, and the face verification task with contrastive loss, which reduces the variation in within-expression features, are combined to train the MSCNN.}
		\label{task}
	\end{figure}
	
	Besides, several works \cite{meng2017identity, zhang2017facial} employed multitask learning for identity-invariant FER. In \cite{meng2017identity}, an identity-aware CNN (IACNN) with two identical sub-CNNs was proposed. One stream used expression-sensitive contrastive loss  to learn expression-discriminative features, and the other stream used identity-sensitive contrastive loss to learn identity-related features for identity-invariant FER.
	In \cite{zhang2017facial}, a multisignal CNN (MSCNN),  which was trained under the supervision of both FER and face verification tasks, was proposed to force the model to focus on expression information (see Fig. \ref{task}). 
	Furthermore, an all-in-one CNN model \cite{ranjan2017all} was proposed to simultaneously solve a diverse set of face analysis tasks including smile detection. The network was first initialized using the weight pre-trained on face recognition, then task-specific sub-networks were branched out from different layers with  domain-based regularization by training on multiple datasets. Specifically, as smile detection is a subject-independent task that relies more on local information available from the lower layers, the authors proposed to fuse the lower convolutional layers to form a generic representation for smile detection. Conventional supervised multitask learning requires training samples labeled for all tasks.  To relax this, \cite{zhang2018From} proposed a  novel attribute propagation method which can leverage the inherent correspondences between facial expression  and other heterogeneous attributes despite the disparate distributions of different datasets.
	
	\subsubsection{Cascaded networks}
	\begin{figure}[tp]
		\includegraphics[width=8.5cm]{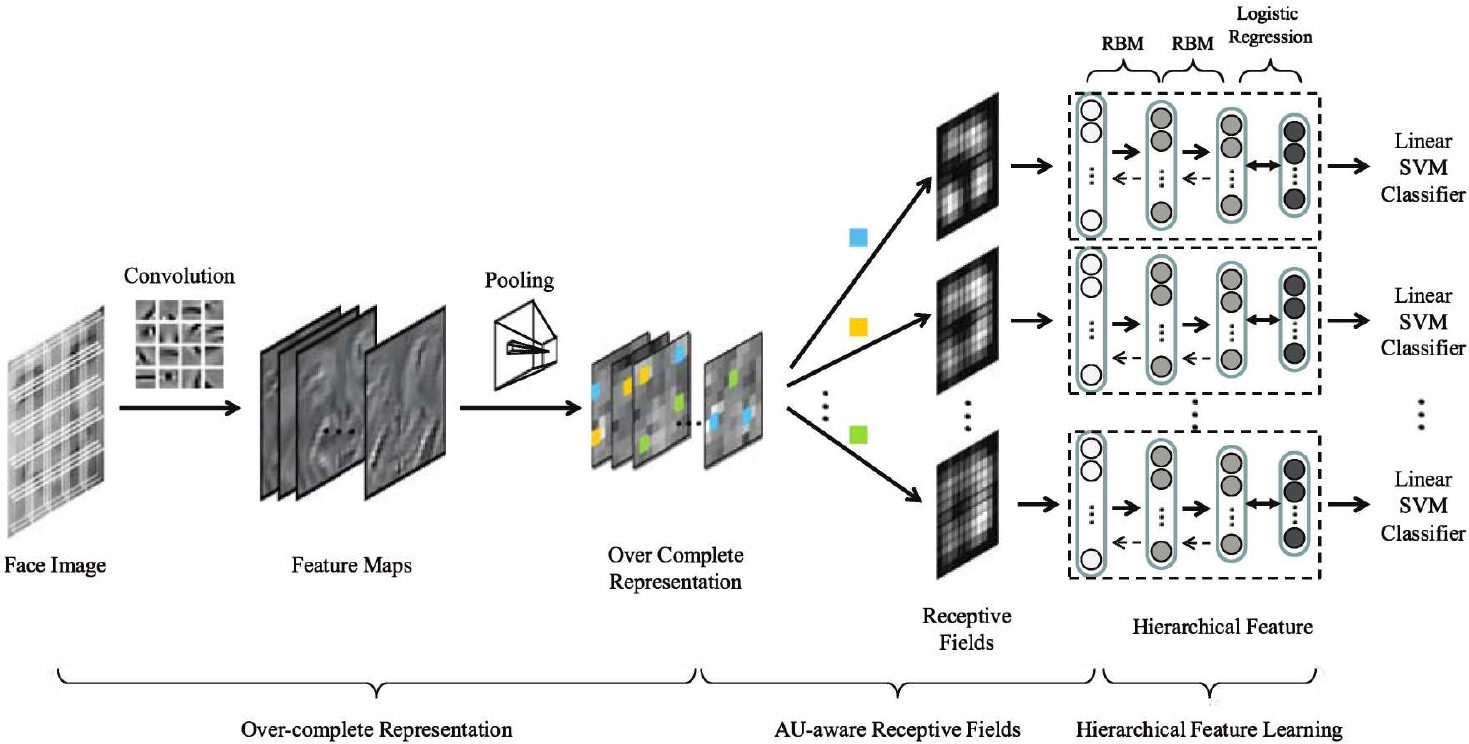}
		\caption{Representative cascaded network for FER. The proposed AU-aware deep network (AUDN) \cite{liu2013aware} is composed of three sequential modules: in the first module, a 2-layer CNN is trained to generate an over-complete representation encoding all expression-specific appearance variations over all possible locations; in the second module, an AU-aware receptive field layer is designed to search subsets of the over-complete representation; in the last module, a multilayer RBM is exploited to learn hierarchical features. }
		\label{cascade}
	\end{figure}
	In a cascaded network, various modules for different tasks are combined sequentially to construct a deeper network, where the outputs of the former modules are utilized by the latter modules.
	Related studies have proposed combinations of different structures to learn a hierarchy of features through which factors of variation that are unrelated with expressions can be gradually filtered out.
	
	Most commonly, different networks or learning methods are combined sequentially and individually, and each of them contributes differently and hierarchically. 
	In \cite{lv2014facial}, DBNs were trained to first detect faces and to detect expression-related areas. Then, these parsed face components were classified by a stacked autoencoder.
	In \cite{rifai2012disentangling}, a multiscale contractive convolutional network (CCNET) was proposed to obtain local-translation-invariant (LTI) representations. Then, contractive autoencoder was designed to hierarchically separate out the emotion-related factors from subject identity and pose.
	In \cite{liu2013aware,liu2015inspired},  over-complete representations were first learned using CNN architecture, then a multilayer RBM was exploited to learn higher-level features for FER (see Fig. \ref{cascade}).
	Instead of simply concatenating different networks, Liu et al. \cite{liu2014facial} presented a boosted DBN (BDBN) that iteratively performed  feature representation, feature selection and classifier construction in a unified loopy state.  Compared with the concatenation without feedback, this loopy framework propagates backward the classification error to initiate the feature selection process alternately until convergence. Thus, the discriminative ability for FER can  be substantially improved during this iteration. 
	
	\subsubsection{Generative adversarial networks (GANs)}
	\label{GAN}
	Recently, GAN-based methods have been successfully used in image synthesis to generate impressively realistic faces, numbers, and a variety of other image types, which are beneficial to training data augmentation and the corresponding recognition tasks. Several works have proposed novel GAN-based models for pose-invariant FER and identity-invariant FER.
	
	For pose-invariant FER, Lai et al. \cite{lai2018emotion} proposed a GAN-based face frontalization framework, where the generator  frontalizes input face images while preserving the identity and expression characteristics and the discriminator distinguishes the real images from the generated frontal face images.
	And Zhang et al. \cite{zhang2018joint} proposed a GAN-based model  that can generate images with different expressions under arbitrary poses for multi-view FER. 
	For identity-invariant FER, Yang et al. \cite{yang2018identity} proposed an Identity-Adaptive  Generation (IA-gen) model with two parts. The upper part generates images of the same subject with different expressions using cGANs, respectively. Then, the lower part conducts FER for each single identity sub-space without involving other individuals, thus identity variations can be well alleviated.
	Chen et al. \cite{chen2018vgan} proposed a Privacy-Preserving
	Representation-Learning Variational GAN (PPRL-VGAN) that combines VAE and GAN to learn an identity-invariant representation that is explicitly disentangled from the identity information and
	generative for expression-preserving face image synthesis.
	Yang et al. \cite{yang2018facial} proposed a De-expression
	Residue Learning (DeRL) procedure to explore the
	expressive information, which is filtered out  during the de-expression process but still embedded in the generator. Then the model extracted this information from the generator directly to mitigate the influence of subject variations and improve the FER performance.  
	
	\subsubsection{Discussion}
	\begin{table*}[t]
		\centering
		\renewcommand{\arraystretch}{1.2}
		\setlength{\tabcolsep}{10pt}
		\caption{Comparison of different types of methods for static images in terms of  data size requirement, variations* (head pose, illumination, occlusion and other environment factors), identity bias, computational efficiency, accuracy,  and difficulty on network training.}
		\label{method_com1}
		\begin{tabular}{@{}|lcccccc|@{}}
			\hline
			Network type	&  data&variations*&identity bias &
			efficiency&accuracy &difficulty  \\\hline
			Pre-train \& Fine-tune &low&fair  &vulnerable  &high  &fair &easy \\\hline
			Diverse input          &low&good  &vulnerable  &low   &fair &easy \\\hline
			Auxiliary layers       &varies&good&varies &varies&good &varies \\\hline
			Network ensemble       &low  &good  &fair  &low   &good &medium \\\hline
			Multitask network      &high&varies  &good &fair   &varies&hard \\\hline
			Cascaded network       &fair&good  &fair &fair  &fair &medium\\\hline
			GAN                    &fair&good  &good &fair  &good &hard  \\\hline
		\end{tabular}
	\end{table*}
	The existing well-constructed deep FER systems focus on two key issues: the lack of plentiful diverse training data and expression-unrelated variations, such as illumination, head pose and identity. Table \ref{method_com1} shows relative
	advantages and disadvantages of these different types of methods with respect to two open issues (data size requirement and expression-unrelated variations) and other focuses (computation efficiency, performance and difficulty of network training).
	
	\textit{ Pre-training and fine-tuning} have become mainstream in deep FER to solve the problem of insufficient training data and overfitting. A practical technique that proved to be particularly useful is pre-training and fine-tuning the network in multiple stages using auxiliary data from large-scale objection or face recognition datasets to small-scale FER datasets, i.e., from large to small and from general to specific. However, when compared with the end-to-end training framework, the representational structure that are unrelated to expressions are still remained in the off-the-shelf pre-trained model, such as the large domain gap with the objection net \cite{kaya2017video} and the subject identification distraction in the face net \cite{ding2017facenet2expnet}. Thus the extracted features are usually vulnerable to identity variations and the performance would be degraded. Noticeably, with the advent of large-scale in-the-wild FER datasets (e.g., AffectNet and RAF-DB), the end-to-end training using deep networks with moderate size can also achieve competitive performances \cite{li2018reliable,zeng2018facial}.
	
	In addition to directly using the raw image data to train the deep network, \textit{diverse pre-designed features} are recommended to strengthen the network's robustness to common distractions (e.g., illumination, head pose and occlusion) and to force the network to focus more on facial areas with expressive information. Moreover, the use of multiple heterogeneous input data can indirectly enlarge the data size. However, the problem of identity bias is commonly ignored in this methods. Moreover, generating diverse data accounts for additional time consuming and combining these multiple data can lead to high dimension which may influence the computational efficiency of the network. 
	
	
	Training a deep and wide network with a large number of hidden layers and flexible filters is  an effective way to learn deep high-level features that are discriminative for the target task. However, this process is vulnerable to the size of training data and can underperform  if insufficient training data is available to learn the new parameters. Integrating multiple relatively small networks in parallel or in series is a natural research direction to overcome this problem.
	\textit{Network ensemble} integrates diverse networks at the feature  or decision level to combine their advantages, which is usually applied in emotion competitions to help boost the performance. However, designing different kinds of networks to compensate each other obviously enlarge the computational cost and the storage requirement.  Moreover, the weight of each sub-network is usually learned according to  the performance on original training data, leading to overfitting on newly unseen testing data.
	\textit{Multitask networks} jointly train multiple networks with consideration of interactions between the target FER task and other secondary tasks, such as facial landmark localization, facial AU recognition and face verification, thus the expression-unrelated factors including identity bias can be well disentangled. The downside of this method is that it requires labeled data from all tasks and the training becomes increasingly cumbersome as more tasks are involved.
	Alternatively, \textit{cascaded networks} sequentially train multiple networks in a hierarchical approach, in which case the discriminative ability of the learned features are continuously strengthened. In general, this method can alleviate the overfitting problem, and in the meanwhile, progressively disentangling factors that are irrelevant to facial expression. A deficiency worths considering is that the sub-networks in most existing cascaded systems are training individually without feedback, and the end-to-end training strategy is preferable to enhance the training effectiveness and the performance \cite{liu2014facial}.
	
	Ideally, deep networks, especially CNNs, have good capabilities for dealing with head-pose variations, yet most current FER networks do not address head-pose variations explicitly and are not tested in naturalistic scenarios. \textit{Generative adversarial networks (GANs)} can be exploited to solve this issue by frontalizing face images while preserving expression characteristics \cite{lai2018emotion} or synthesizing arbitrary poses to help train the pose-invariant network \cite{zhang2018joint}. Another advantage of GANs is that the identity variations can be explicitly disentangled through generating the corresponding neutral face image \cite{yang2018facial} or synthesizing different expressions while preserving the identity information for identity-invariant FER \cite{yang2018identity}. Moreover, GANs can help augment the training data on both size and diversity. The main drawback of GAN is the training instability  and the trade-off between visual quality and image diversity.

	\subsection{Deep FER networks for dynamic image sequences}
	
	Although most of the previous models focus on static images, facial expression recognition can benefit from the temporal correlations of consecutive frames in a sequence. We first introduce the existing frame aggregation techniques that strategically combine deep features learned from static-based FER networks. Then, considering that in a videostream people usually display the same expression with different intensities, we further review methods that use images in different expression intensity states for intensity-invariant FER.
	Finally, we introduce deep FER networks that consider spatio-temporal motion patterns in video frames and learned features derived from the temporal structure.
	For each of the most frequently evaluated datasets, Table \ref{result2} shows the current state-of-the-art methods  conducted in the person-independent protocol.
	
	\begin{table*}[!t]
		\scriptsize
		\renewcommand{\arraystretch}{1}
		\setlength{\tabcolsep}{1.1pt}
		\centering
		\caption{Performances of representative methods for dynamic-based deep facial expression recognition on the most widely evaluated datasets. Network size = depth \& number of parameters; Pre-processing = Face Detection \& Data Augmentation \& Face Normalization; IN = Illumination Normalization; $\mathcal{FA}$ = Frame Aggregation; $\mathcal{EIN}$ = Expression Intensity-invariant Network;  $\mathcal{FLT}$ = Facial Landmark Trajectory; $\mathcal{CN}$ = Cascaded Network; $\mathcal{NE}$ = Network Ensemble; S = Spatial Network; T = Temporal Network; LOSO = leave-one-subject-out.}
		\label{result2}
		\begin{threeparttable}
			\begin{tabular}{@{}
					|m{0.05\textwidth}<{\centering}
					|m{0.14\textwidth}<{\centering}
					|m{0.08\textwidth}<{\centering}
					|m{0.03\textwidth}<{\centering}|c
					|c|c|c
					|c
					|c
					|m{0.06\textwidth}<{\centering}
					|c|@{}}
				\hline
				Datasets 
				& Methods 
				&\begin{tabular}[c]{@{}c@{}} Network\\ type\end{tabular}
				&\multicolumn{2}{c|}{\begin{tabular}[c]{@{}c@{}} Network\\ size\end{tabular}}
				&\multicolumn{3}{c|}{Pre-processing}
				&\begin{tabular}[c]{@{}c@{}}Training data Selection\\in each sequence\end{tabular}
				&\begin{tabular}[c]{@{}c@{}}Testing data selection\\in each sequence\end{tabular}
				& Data group 
				& Performance\tnote{1} (\%) \\ \hline\hline
				
				\multirow{12}{*}{\textbf{CK+}}
				
				&Zhao et al. 16 \cite{zhao2016peak}
				&$\mathcal{EIN}$
				&22&6.8m
				&\ding{51}&-&-
				&from the 7th to the last\tnote{2}
				&the last frame
				&10 folds
				&6 classes: 99.3\\\cline{2-12}
				
				&Yu et al. 17 \cite{yu2017deeper}
				&$\mathcal{EIN}$
				&42&-
				&MTCNN&\ding{51}&-
				&from the 7th to the last\tnote{2}
				&the peak expression
				&10 folds
				&6 classes: 99.6\\\cline{2-12}
				
				&kim et al. 17 \cite{kim2017deep}
				&$\mathcal{EIN}$
				&14&-
				&\ding{51}&\ding{51}&-
				&all frames
				&\multirow{9}{*}{\begin{tabular}[c]{@{}c@{}}the same as \\the training data\end{tabular}}
				&10 folds
				&7 classes: 97.93 \\\cline{2-9}\cline{11-12}

				&Sun et al. 17 \cite{sun2017deep}
				&$\mathcal{NE}$
				&\multicolumn{2}{c|}{3 * GoogLeNetv2}
				&\ding{51}&-&-
				&\begin{tabular}[c]{@{}c@{}} S: emotional  \\T: neutral$+$emotional \end{tabular}
				&
				&10 folds
				&6 classes: 97.28\\\cline{2-9}\cline{11-12}
				
				&Jung et al. 15\cite{jung2015joint}
				&$\mathcal{FLT}$
				&2&177.6k
				&IntraFace&\ding{51}&-
				&fixed number of frames
				&
				&10 folds
				&7 classes: 92.35\\\cline{2-9}\cline{11-12}
				
				&Jung et al. 15\cite{jung2015joint}
				&C3D
				&4&-
				&IntraFace&\ding{51}&-
				&fixed number of frames
				&
				&10 folds
				&7 classes: 91.44\\\cline{2-9}\cline{11-12}
				
				&Jung et al. 15\cite{jung2015joint}
				&$\mathcal{NE}$
				&\multicolumn{2}{c|}{$\mathcal{FLT}$/C3D}
				&IntraFace&\ding{51}&-
				&fixed number of frames
				&
				&10 folds
				&7 classes: 97.25  (95.22)\\\cline{2-9}\cline{11-12}
				
				&kuo et al. 18 \cite{kuo2018compact}
				&$\mathcal{FA}$
				&6 &2.7m
				&IntraFace&\ding{51}&IN
				&fixed length 9
				&
				&10 folds
				&7 classes: 98.47\\\cline{2-9}\cline{11-12}
				
				&Zhang et al. 17 \cite{zhang2017facial}
				&$\mathcal{NE}$
				&7/5&2k/1.6m
				&\begin{tabular}[c]{@{}c@{}}SDM/\\Cascaded CNN\end{tabular}&\ding{51}&-
				&\begin{tabular}[c]{@{}c@{}}S: the last frame \\T: all frames\end{tabular}
				&
				&10 folds
				&7 classes: 98.50 (97.78)\\\hline\hline

				\multirow{10}{*}{\textbf{MMI}}
				
				&Kim et al. 17 \cite{kim2017multi}
				&$\mathcal{EIN}, \mathcal{CN}$
				&7&1.5m
				&Incremental&\ding{51}&-
				&5 intensities frames
				&\multirow{8}{*}{\begin{tabular}[c]{@{}c@{}}the same as \\the training data\end{tabular}} 
				&LOSO
				&6 classes: 78.61 (78.00)\\\cline{2-9}\cline{11-12}
				
				&kim et al. 17 \cite{kim2017deep}
				&$\mathcal{EIN}$
				&14&-
				&\ding{51}&\ding{51}&-
				&all frames
				&
				&10 folds
				&6 classes: 81.53 \\\cline{2-9}\cline{11-12}
				
				&Hasani et al. 17 \cite{hasani2017facial}
				&
				$\mathcal{FLT, CN}$
				&22&-
				&3000 fps&-&-
				&ten frames
				&
				&5 folds
				&6 classes: 77.50 (74.50)\\\cline{2-9}\cline{11-12}
				
				&Hasani et al. 17 \cite{hasani2017spatio}
				&$\mathcal{CN}$
				&29&-
				&AAM&-&-
				&static frames
				&
				&5 folds
				&6 classes: 78.68\\\cline{2-9}\cline{11-12}
				
				&Zhang et al. 17 \cite{zhang2017facial}
				&$\mathcal{NE}$
				&7/5&2k/1.6m
				&\begin{tabular}[c]{@{}c@{}}SDM/\\Cascaded CNN\end{tabular}&\ding{51}&-
				&\begin{tabular}[c]{@{}c@{}} S: the middle frame\\T: all frames\end{tabular}
				&
				&10 folds
				&6 classes: 81.18 (79.30)\\\cline{2-9}\cline{11-12}
				
				&Sun et al. 17 \cite{sun2017deep}
				&$\mathcal{NE}$
				&\multicolumn{2}{c|}{3 * GoogLeNetv2}
				&\ding{51}&-&-
				&\begin{tabular}[c]{@{}c@{}} S: emotional  \\T: neutral$+$emotional \end{tabular}
				&&10 folds
				&6 classes: 91.46\\\hline\hline

				\multirow{8}{*}{\textbf{\begin{tabular}[c]{@{}c@{}} Oulu-\\CASIA\end{tabular}}}
				
				&Zhao et al. 16 \cite{zhao2016peak}
				&$\mathcal{EIN}$
				&22&6.8m
				&\ding{51}&-&-
				&from the 7th to the last\tnote{2}
				&the last frame
				&10 folds
				&6 classes: 84.59\\\cline{2-12}
				
				&Yu et al. 17 \cite{yu2017deeper}
				&$\mathcal{EIN}$
				&42&-
				&MTCNN&\ding{51}&-
				&from the 7th to the last\tnote{2}
				&the peak expression
				&10 folds
				&6 classes: 86.23\\\cline{2-12}
				
				&Jung et al. 15\cite{jung2015joint}
				&$\mathcal{FLT}$
				&2&177.6k
				&IntraFace&\ding{51}&-
				&fixed number of frames
				&\multirow{6}{*}{\begin{tabular}[c]{@{}c@{}}the same as \\the training data\end{tabular}}
				&10 folds
				&6 classes: 74.17\\\cline{2-9}\cline{11-12}
				
				&Jung et al. 15\cite{jung2015joint}
				&C3D
				&4&-
				&IntraFace&\ding{51}&-
				&fixed number of frames
				&
				&10 folds
				&6 classes: 74.38\\\cline{2-9}\cline{11-12}
				
				&Jung et al. 15 \cite{jung2015joint}
				&$\mathcal{NE}$
				&\multicolumn{2}{c|}{$\mathcal{FLT}$/C3D}
				&IntraFace&\ding{51}&-
				&fixed number of frames
				&
				&10 folds
				&6 classes: 81.46 (81.49)\\\cline{2-9}\cline{11-12}
				
				&Zhang et al. 17 \cite{zhang2017facial}
				&$\mathcal{NE}$
				&7/5&2k/1.6m
				&\begin{tabular}[c]{@{}c@{}}SDM/\\Cascaded CNN\end{tabular}&\ding{51}&-
				&\begin{tabular}[c]{@{}c@{}} S: the last frame\\T: all frames\end{tabular}
				&
				&10 folds
				&6 classes: 86.25 (86.25)\\\cline{2-9}\cline{11-12}
				
				&kuo et al. 18 \cite{kuo2018compact}
				&$\mathcal{NE}$
				&6 &2.7m
				&IntraFace&\ding{51}&IN
				&fixed length 9
				&
				&10 folds
				&6 classes: 91.67\\\hline\hline
				
				\multirow{7}{*}{\textbf{\begin{tabular}[c]{@{}c@{}}AFEW\tnote{*} \\6.0\end{tabular}}}
				
				&Ding et al. 16 \cite{ding2016audio}
				&$\mathcal{FA}$
				&\multicolumn{2}{c|}{AlexNet}
				&\ding{51}&-&-
				&\multicolumn{3}{c|}{Training: 773; Validation: 373; Test: 593} 
				&Validation: 44.47\\\cline{2-12}
				
				&Yan et al. 16 \cite{yan2016multi}
				&$\mathcal{CN}$
				&\multicolumn{2}{c|}{VGG16-LSTM}
				&\ding{51}&\ding{51}&-
				&\multicolumn{2}{c|}{40 frames}
				&3 folds
				&7 classes: 44.46\\\cline{2-12}
				
				&Yan et al. 16 \cite{yan2016multi}
				&$\mathcal{FLT}$
				&4&-
				&\cite{cui2018recurrent}&-&-
				&\multicolumn{2}{c|}{30 frames}
				&3 folds
				&7 classes: 37.37\\\cline{2-12}
				
				&Fan et al. 16 \cite{fan2016video}
				&$\mathcal{CN}$
				&\multicolumn{2}{c|}{VGG16-LSTM}
				&\ding{51}&-&-
				&\multicolumn{3}{c|}{16 features for LSTM}
				&Validation: 45.43 (38.96)\\\cline{2-12}
				
				&Fan et al.  \cite{fan2016video}
				&C3D
				&10&-
				&\ding{51}&-&-
				&\multicolumn{3}{c|}{several windows of 16 consecutive frames}
				&Validation: 39.69 (38.55)\\\cline{2-12}
				
				&Yan et al. 16 \cite{yan2016multi}
				&fusion
				&\multicolumn{5}{c|}{/}
				&\multicolumn{3}{c|}{Training: 773; Validation: 383; Test: 593} &Test: 56.66 (40.81)\\\cline{2-12}
				
				&Fan et al. 16 \cite{fan2016video}
				&fusion
				&\multicolumn{5}{c|}{/}
				&\multicolumn{3}{c|}{Training: 774; Validation: 383; Test: 593}
				&Test: 59.02 (44.94)\\\hline

				\multirow{5}{*}{\textbf{\begin{tabular}[c]{@{}c@{}}AFEW\tnote{*} \\7.0\end{tabular}}}
				&Ouyang et al. 17 \cite{ouyang2017audio}
				&$\mathcal{CN}$
				&\multicolumn{2}{c|}{VGG-LSTM}
				&MTCNN&\ding{51}&-
				&\multicolumn{3}{c|}{16 frames}
				&Validation: 47.4\\\cline{2-12}
				
				&Ouyang et al. 17 \cite{ouyang2017audio}
				&C3D
				&10&-
				&MTCNN&\ding{51}&-
				&\multicolumn{3}{c|}{16 frames}
				&Validation: 35.2\\\cline{2-12}
				
				&Vielzeuf et al.  \cite{vielzeuf2017temporal} 
				&$\mathcal{CN}$
				&\multicolumn{2}{c|}{C3D-LSTM}
				&\ding{51}&\ding{51}&-
				&\multicolumn{3}{c|}{detected face frames}
				&Validation: 43.2\\\cline{2-12}
				
				&Vielzeuf et al. \cite{vielzeuf2017temporal} 
				&$\mathcal{CN}$
				&\multicolumn{2}{c|}{VGG16-LSTM}
				&\ding{51}&\ding{51}&-
				&\multicolumn{3}{c|}{several windows of 16 consecutive frames}&Validation: 48.6\\\cline{2-12}
				
				&Vielzeuf et al.  \cite{vielzeuf2017temporal} 
				&fusion
				&\multicolumn{5}{c|}{/}
				&\multicolumn{3}{c|}{\multirow{1}{*}{Training: 773; Validation: 383; Test: 653}}
				&Test: 58.81 (43.23)\\\hline

			\end{tabular}
			\begin{tablenotes}
				\item[1] The value in parentheses is the mean accuracy calculated from the confusion matrix given by authors.
				\item[2] A pair of images (peak and non-peak expression) is chosen for training each time.
				\item[*] We have included the result of a single spatio-temporal network and also the best result after fusion with both video and audio modalities.
				\item[$\dagger$] 7 Classes in CK+: Anger, Contempt, Disgust, Fear, Happiness, Sadness, and Surprise.
				\item[$\ddagger$] 7 Classes in AFEW: Anger, Disgust, Fear, Happiness, Neutral, Sadness, and Surprise.
			\end{tablenotes}
		\end{threeparttable}
	\end{table*}
	
	\subsubsection{Frame aggregation}
	Because the frames in a given video clip may vary in expression intensity, directly measuring per-frame error does not yield satisfactory performance. Various methods have been proposed to aggregate the network output for frames in each sequence to improve the performance. We divide these methods into two groups: decision-level frame aggregation and feature-level frame aggregation.
	
\begin{figure}[t]
		\subfigure[Frame averaging]{\label{frame1}
			\includegraphics[width=4cm]{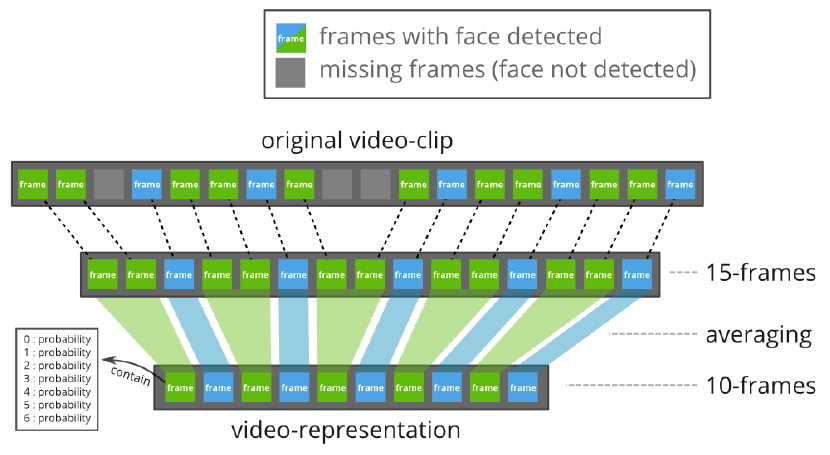}}
		\subfigure[Frame expansion]{\label{frame2}
			\includegraphics[width=4cm]{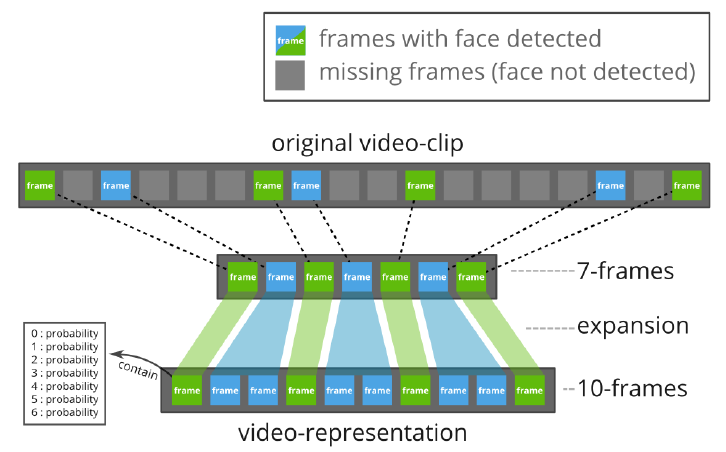}}
		\caption{Frame aggregation in \cite{kahou2013combining}. The flowchart is top-down. (a) For sequences with more than 10 frames, we averaged the probability vectors of 10 independent groups of frames taken uniformly along time.  (b)  For sequences with less than 10 frames, we expanded by repeating frames uniformly to obtain 10 total frames.}
		\label{frame}
	\end{figure}

	For decision-level frame aggregation,  $n$-class probability vectors of each frame in a sequence are integrated. The most convenient way is to directly concatenate the output of these frames. However, the number of frames in each sequence may be different. Two aggregation approaches have been considered to generate a fixed-length feature vector for each sequence \cite{kahou2013combining,kahou2016emonets}: frame averaging and frame expansion (see Fig. \ref{frame} for details). An alternative approach which dose not require a fixed number of frames is applying statistical coding. The average, max, average of square, average of maximum suppression vectors and so on can be used to summarize the per-frame probabilities in each sequence.
	
	For feature-level frame aggregation, the learned features of frames in the sequence are aggregate.  Many statistical-based encoding modules can be applied in this scheme. A simple and effective way is to concatenate the mean, variance, minimum, and maximum of the features over all frames \cite{bargal2016emotion}. 
	Alternatively, matrix-based models such as eigenvector, covariance matrix and multi-dimensional Gaussian distribution can also be employed for aggregation  \cite{liu2014combining, ding2016audio}. 
	Besides, multi-instance learning has been explored for video-level representation \cite{xu2016video}, where the cluster centers are computed from auxiliary image data and then bag-of-words representation is obtained for each bag of video frames.

	\subsubsection{ Expression Intensity network}
	\begin{figure}[t]
		\includegraphics[width=8.5cm]{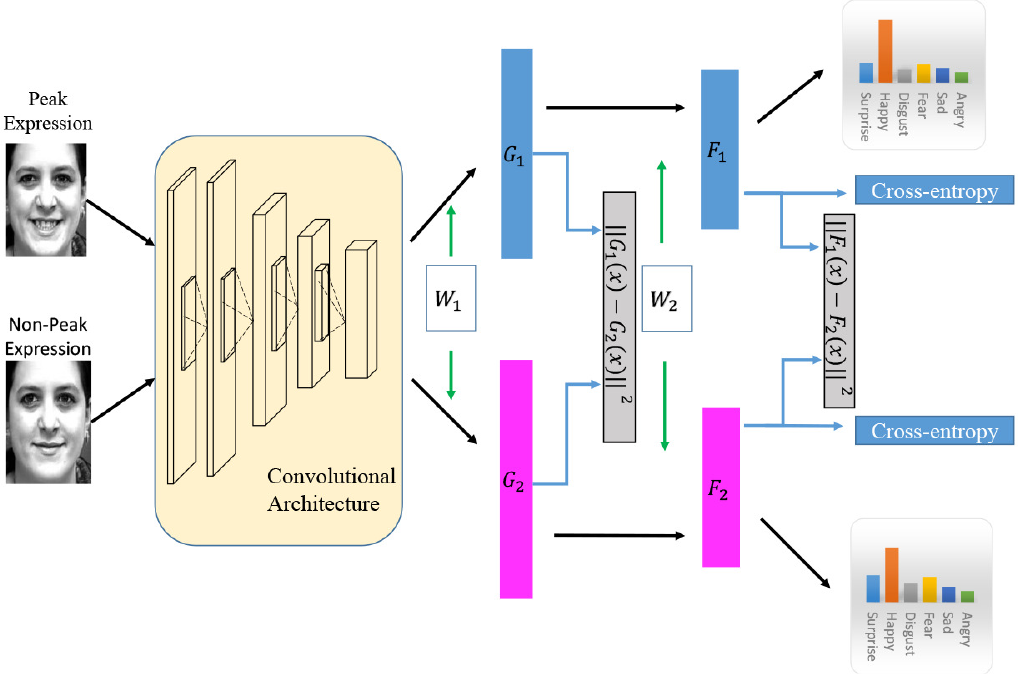}
		\caption{The proposed PPDN in \cite{zhao2016peak}. During training, PPDN is trained  by jointly optimizing the L2-norm loss and the cross-entropy losses of  two expression images. During testing, the PPDN takes one still image as input for probability prediction.}
		\label{intensity}
	\end{figure}
	Most methods (introduced in Section \ref{sta}) focus on recognizing the peak high-intensity expression and ignore the subtle lower-intensity expressions. In this section, we introduced expression intensity-invariant networks that take training samples with different intensities as input to exploit the intrinsic correlations among expressions from a sequence that vary in intensity.
	
	In expression intensity-invariant network, image frames with intensity labels are used for training. During test, data that vary in expression intensity are used to verify the  intensity-invariant ability of the network. 
	Zhao et al. \cite{zhao2016peak} proposed a peak-piloted deep network (PPDN) that takes a pair of peak and non-peak images of the same expression and from the same subject as input and utilizes the L2-norm loss to minimize the distance between both images. During  back propagation, a peak gradient suppression (PGS) was proposed to drive the learned feature of the non-peak expression towards that of peak expression while avoiding the inverse. Thus, the network discriminant ability on lower-intensity expressions can be improved.
	Based on PPDN, Yu et al. \cite{yu2017deeper} proposed a deeper cascaded peak-piloted network (DCPN) that used a deeper and larger  architecture to enhance the discriminative ability of the learned features and employed an integration training method called cascade fine-tuning to avoid over-fitting.
	In \cite{kim2017multi}, more intensity states were utilized (onset, onset to apex transition, apex, apex to offset transition and offset) and five loss functions were adopted to regulate the network training by minimizing expression classification error, intra-class expression variation, intensity classification error and intra-intensity variation, and encoding intermediate intensity, respectively.

	Considering that images with different expression intensities for an individual identity is not always available in the wild, 
	several works proposed to automatically acquire the intensity label or to generate new images with targeted intensity. For example, in \cite{chen2018deep} the peak and neutral frames was automatically picked out from the sequence with two stages : a clustering stage to divide all frames into the peak-like group and the neutral-like group using K-means algorithm, and a classification stage to detect peak and neutral frames using a semi-supervised SVM.
	And in \cite{kim2017deep}, a deep generative-contrastive model was presented  with two steps: a generator to generate the reference (less-expressive) face for each sample via convolutional encoder-decoder  and a contrastive network to jointly filter out information that is irrelevant with expressions through a contrastive metric loss and a supervised reconstruction loss. 
	
	\subsubsection{Deep spatio-temporal  FER network}
	Although the frame aggregation can integrate frames in the video sequence, the  crucial temporal dependency is not explicitly exploited.
	By contrast, the spatio-temporal FER network takes a range of frames in a temporal window as a single input without prior knowledge of the expression intensity and utilizes both textural  and temporal information to encode more subtle expressions.
	\\\\
	\textbf{RNN and C3D:}
	\label{dtan}
	RNN can robustly derive information from sequences  by exploiting the fact that feature vectors for successive data are connected semantically and are therefore interdependent. The improved version, LSTM, is flexible to handle varying-length sequential data with lower computation cost.
	Derived from RNN, an RNN that is composed of ReLUs and initialized with the identity matrix (IRNN) \cite{le2015simple} was used to provide a simpler mechanism for addressing the vanishing and exploding gradient problems \cite{ebrahimi2015recurrent}. And bidirectional RNNs (BRNNs) \cite{schuster1997bidirectional} were employed to learn the temporal relations in both the original and reversed directions \cite{yan2016multi,zhang2017facial}.
	Recently, a Nested LSTM was proposed in \cite{yu2018spatio} with two sub-LSTMs. Namely, T-LSTM models the temporal dynamics of the learned features, and C-LSTM integrates the outputs of all T-LSTMs together so as to encode the multi-level features encoded in the intermediate layers of the network.
	
\begin{figure}[t]
		\includegraphics[width=8.5cm]{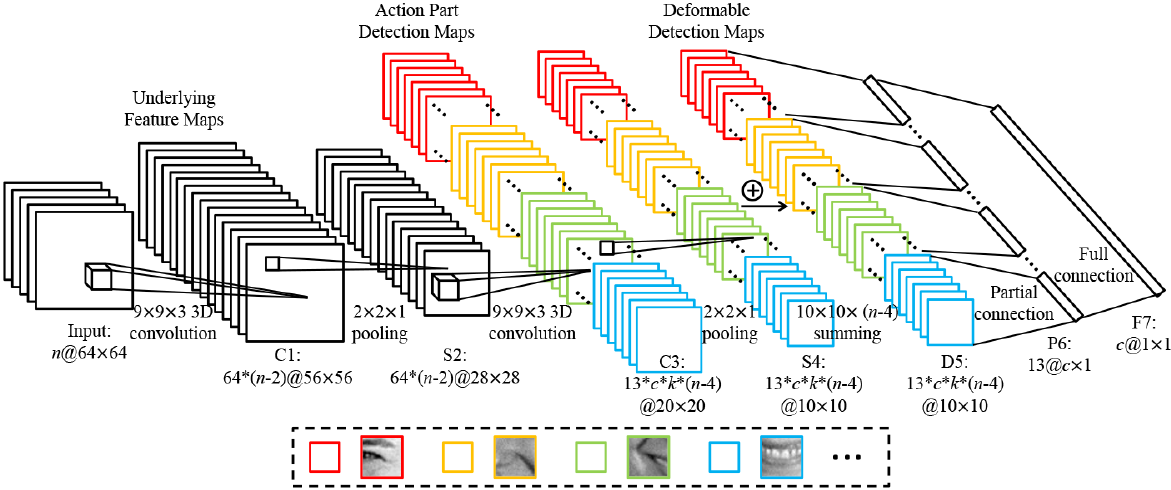}
		\caption{The proposed 3DCNN-DAP \cite{liu2014deeply}. The input $n$-frame sequence is convolved with 3D filters; then, $13*c*k$ part filters corresponding to $13$ manually defined facial parts are used to convolve $k$ feature maps for the facial action part detection maps of $c$ expression classes.}
		\label{c3d}
	\end{figure}

	Compared with RNN, CNN is more suitable for computer vision applications; hence, its derivative C3D \cite{tran2015learning}, which uses 3D convolutional kernels with shared weights along the time axis instead of the traditional 2D kernels, has been widely used for dynamic-based FER (e.g., \cite{barros2016Developing,fan2016video,ouyang2017audio,abbasnejad2017using,zhao2018learning}) to capture the spatio-temporal features. 
	Based on C3D, many derived structures have been designed for FER. 
	In \cite{liu2014deeply}, 3D CNN was incorporated with the DPM-inspired \cite{felzenszwalb2010object} deformable facial action constraints to simultaneously encode dynamic motion and discriminative part-based representations (see Fig.
	\ref{c3d} for details).
	In \cite{jung2015joint}, a deep temporal appearance network (DTAN) was proposed that employed 3D filters without weight sharing along the time axis; hence, each filter can vary in importance over time. 
	Likewise, a weighted C3D was proposed \cite{vielzeuf2017temporal}, where several windows of consecutive frames were extracted from each sequence and weighted based on their prediction scores.
	Instead of directly using C3D for classification, \cite{nguyen2017deep} employed C3D for spatio-temporal feature extraction and then cascaded with DBN for prediction. 
	In \cite{pini2017modeling}, C3D was also used as a feature extractor, followed by a NetVLAD layer \cite{arandjelovic2016netvlad} to aggregate the temporal information of the motion features by learning cluster centers.
	\\\\
	\textbf{Facial landmark trajectory:}
	Related psychological studies have shown that expressions are invoked by dynamic motions of certain facial parts (e.g., eyes, nose and mouth) that contain the most descriptive information for representing expressions. To obtain more accurate facial actions for FER, facial landmark trajectory models have been proposed to capture the dynamic variations of facial components from consecutive frames.
	
\begin{figure}[t]
		\includegraphics[width=8.5cm]{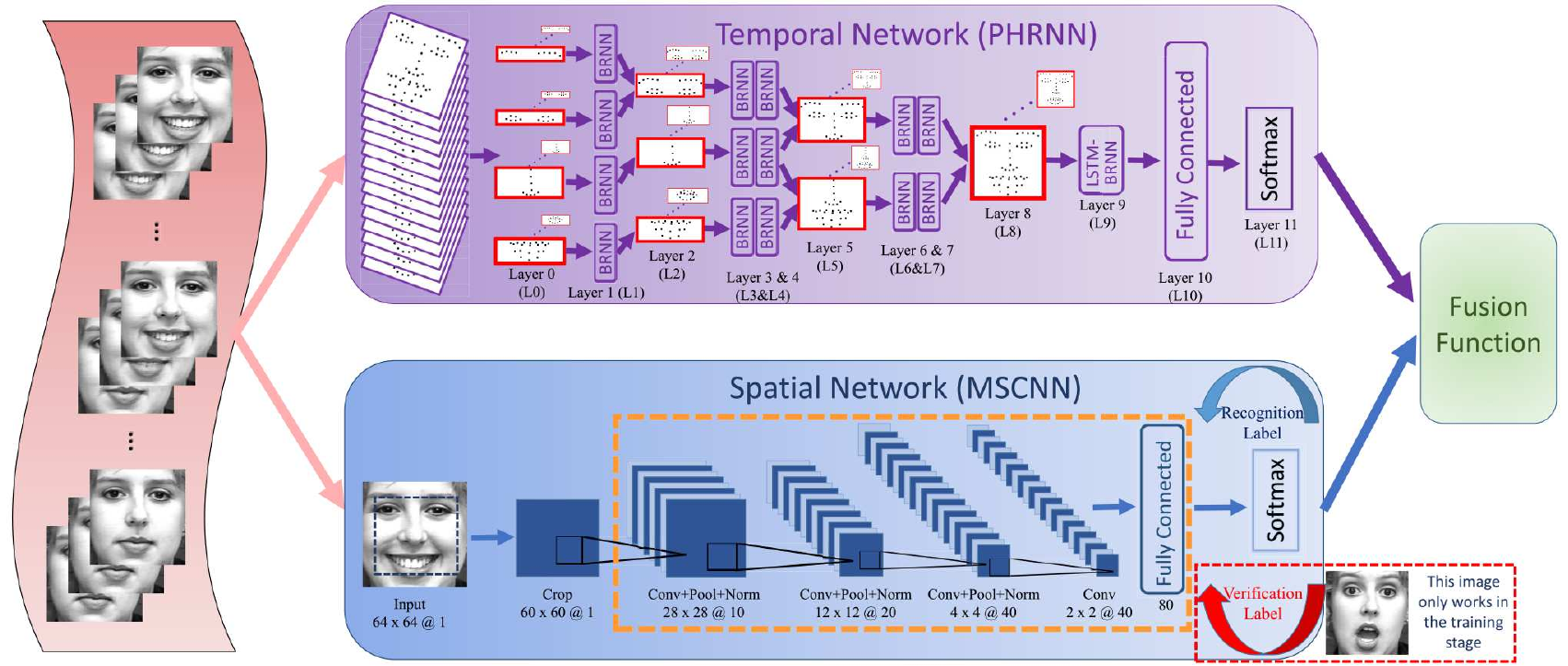}
		\caption{The spatio-temporal network proposed in \cite{zhang2017facial}. The temporal network PHRNN  for landmark trajectory and the spatial network MSCNN for identity-invariant features are trained separately. Then, the predicted probabilities from the two networks are fused together for spatio-temporal FER.}
		\label{stream1}
	\end{figure}

	To extract landmark trajectory representation, the most direct way is to concatenate coordinates of facial landmark points from frames over time with normalization to generate a one-dimensional trajectory signal for each sequence \cite{jung2015joint} or to form an image-like map  as the input of CNN \cite{yan2016multi}.
	Besides, relative distance variation of each landmark in consecutive frames can also be used to capture the temporal information \cite{kim2017multi2}.
	Further, part-based model that divides facial landmarks into several parts according to facial physical structure and then separately feeds them into the networks hierarchically is proved to be efficient for both local low-level and global high-level feature encoding \cite{zhang2017facial} (see ``PHRNN'' in Fig. \ref{stream1}) .
	Instead of separately extracting the trajectory features and then input them into the networks, Hasani et al.  \cite{hasani2017facial} incorporated the trajectory features  by replacing the shortcut in the residual unit of the original 3D Inception-ResNet with element-wise multiplication of facial landmarks and the input tensor of the residual unit. Thus, the landmark based network can be trained end-to-end.
	\\\\
	\textbf{Cascaded networks:}
	By combining the powerful  perceptual vision representations learned from CNNs with the strength of LSTM for variable-length inputs and outputs,
	Donahue et al. \cite{donahue2015long} proposed a both spatially and temporally deep model 
	which cascades the outputs of CNNs with LSTMs for  various vision tasks involving time-varying inputs and outputs. Similar to this hybrid network, many cascaded networks have been proposed for FER (e.g., \cite{fan2016video,kim2017multi,jain2017multi,vielzeuf2017temporal}).
	
	Instead of CNN, \cite{baccouche2012spatio} employed a convolutional sparse autoencoder for sparse and shift-invariant features; then, an LSTM classifier was trained for temporal evolution.
	\cite{ouyang2017audio} employed  a more flexible network called  ResNet-LSTM, which allows nodes in lower CNN layers to directly contact with LSTMs to capture spatio-temporal information.
	In addition to concatenating LSTM with the fully connected layer of CNN, a hypercolumn-based system \cite{kankanamge2017facial} extracted the last convolutional layer features as the input of the LSTM for longer range dependencies without losing global coherence.
	Instead of LSTM, the conditional random fields (CRFs) model \cite{lafferty2001conditional} that are effective in recognizing human activities
	was employed in \cite{hasani2017spatio} to distinguish the temporal relations of the input sequences.
	\\\\
	\textbf{Network ensemble:}
	\begin{figure}[t]
		\setlength{\abovecaptionskip}{2pt}
		\includegraphics[width=8.5cm]{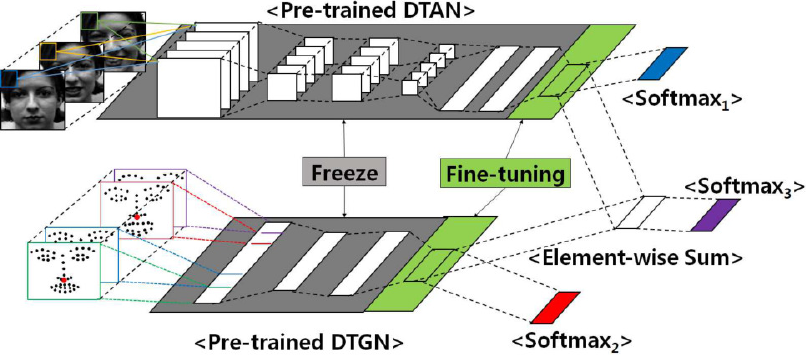}
		\caption{The joint fine-tuning method for DTAGN proposed in \cite{jung2015joint}.
			To integrate DTGA and DTAN, we freeze the weight values in the gray boxes  and retrain the top layer in the green boxes. The logit values of the green boxes are used by Softmax3 to supervise the integrated network. During training, we combine three softmax loss functions, and for prediction, we use only Softmax3.}
		\label{stream2}
	\end{figure}
	A two-stream CNN  for action recognition in videos, which trained one stream of the CNN on the multi-frame dense optical flow for temporal information and the other stream of the CNN on still images for appearance features and then fused the outputs of two streams, was introduced by Simonyan et al. \cite{simonyan2014two}. Inspired by this architecture, several network ensemble models have been proposed for FER. 
	
	Sun et al. \cite{sun2017deep} proposed a multi-channel network that extracted  the spatial information from emotion-expressing faces  and temporal information (optical flow) from the changes between emotioanl and neutral faces, and investigated three feature fusion strategies: score average fusion, SVM-based fusion and neural-network-based fusion. 
	Zhang et al. \cite{zhang2017facial} fused the temporal network PHRNN (discussed in \textbf{``Landmark trajectory''}) and the spatial network MSCNN (discussed in section \ref{mscnn}) to extract the partial-whole, geometry-appearance, and static-dynamic information for FER (see Fig. \ref{stream1}). 
	Instead of fusing the network outputs with different weights, Jung et al. \cite{jung2015joint} proposed a joint fine-tuning method that jointly trained the DTAN (discussed in the \textbf{``RNN and C3D'' }), the DTGN (discussed in the \textbf{``Landmark trajectory''}) and the integrated network (see Fig. \ref{stream2} for details), which outperformed the weighed sum strategy.
	
	\subsubsection{Discussion}
	\begin{table*}[t]
		\centering
		\renewcommand{\arraystretch}{1.2}
		\setlength{\tabcolsep}{10pt}
		\caption{Comparison of different types of methods for dynamic image sequences in terms of data size requirement, representability of spatial and temporal information, requirement on frame length, performance,  and computational efficiency.  $\mathcal{FLT}$ = Facial Landmark Trajectory; $\mathcal{CN}$ = Cascaded Network; $\mathcal{NE}$ = Network Ensemble.}
		\label{method_com2}
		\begin{tabular}{@{}|m{0.1\textwidth}|l||cccccc|@{}}
			\hline
			\multicolumn{2}{|c|}{Network type}   & data&spatial& temporal&frame length&accuracy &efficiency  \\\hline
			\multicolumn{2}{|l||}{Frame aggregation} &low&good  &no  &depends  &fair &high \\\hline
			\multicolumn{2}{|l||}{Expression intensity}&fair&good  &low  &fixed  &fair &varies \\\hline
			\multirow{5}{*}{\begin{tabular}[c]{@{}c@{}} Spatio-\\temporal\\ network\end{tabular}}
			&RNN    &low&low&good &variable&low &fair \\\cline{2-8}
			&C3D    &high  &good  &fair  &fixed   &low &fair \\\cline{2-8}
			&$\mathcal{FLT}$     &fair&fair  &fair &fixed  &low &high \\\cline{2-8}
			&$\mathcal{CN}$      &high&good  &good &variable   &good&fair \\\cline{2-8}
			&$\mathcal{NE}$     &low&good  &good &fixed  &good &low\\\hline
			
		\end{tabular}
	\end{table*}
	In the real world, people display facial expressions in a dynamic process, e.g., from subtle to obvious, and it has become a trend to conduct FER on sequence/video data. 
	Table \ref{method_com2} summarizes relative merits of different types of methods on dynamic data in regards to the capability of representing spatial and temporal information, the requirement on training data size and frame length (variable or fixed), the computational efficiency and the performance. 
	
	\textit{Frame aggregation} is employed to  combine the learned feature or prediction probability of each frame for a sequence-level result. The output of each frame can be simply concatenated (fixed-length frames is required in each sequence) or statistically
	aggregated to obtain video-level representation (variable-length frames processible). This method is computationally simple and  can achieve moderate performance if the temporal variations of the target dataset is not complicated.
	
	According to the fact that the expression intensity in a video sequence varies over time, the \textit{expression intensity-invariant network}  considers images with non-peak expressions and further exploits the dynamic correlations between peak and non-peak expressions to improve performance. Commonly, image frames with specific intensity states are needed for intensity-invariant FER.
	
	Despite the advantages of these methods, \textit{frame aggregation} handles frames  without consideration of temporal information and subtle appearance changes, and \textit{expression intensity-invariant networks} require prior knowledge of expression intensity which is unavailable in real-world scenarios. By contrast, \textit{Deep spatio-temporal networks} are designed to encode temporal dependencies in consecutive frames and have been shown to benefit from learning spatial features in conjunction with temporal features. \textit{RNN and its variations (e.g., LSTM, IRNN and BRNN) and C3D} are foundational networks for learning spatio-temporal features. However, the performance of these networks is barely satisfactory.  RNN is incapable of capturing the powerful convolutional features. And 3D filers in C3D are applied over very short video clips 
	ignoring long-range dynamics. Also, training such a huge network is computationally a problem, especially for dynamic FER where video data is insufficient.
	Alternatively, \textit{facial landmark trajectory} methods extract shape features based on the physical structures of facial morphological variations to capture  dynamic facial component activities, and then apply deep networks for classification. This method is computationally simple and can get rid of the issue on illumination variations. However, it is sensitive to registration errors and requires accurate facial landmark detection, which is difficult to access in unconstrained conditions. Consequently, this method performs less well and is more suitable to complement appearance representations.  
	\textit{Network ensemble} is utilized to train multiple networks for both spatial and temporal information and then to fuse the network outputs in the final stage. Optic flow and facial landmark trajectory can be used as temporal representations to collaborate spatial representations. One of the drawbacks of this framework is the pre-computing and storage consumption on optical flow or landmark trajectory vectors. 
	And most related researches randomly selected fixed-length video frames as input, leading to the loss of useful temporal information.
	\textit{Cascaded networks} were proposed to
	first extract discriminative representations for facial expression images and then input these features to sequential networks to reinforce the temporal information encoding. However, this model introduces additional parameters to capture sequence information, and the feature learning network (e.g., CNN) and the temporal information encoding network (e.g., LSTM) in current works are not trained jointly, which may lead to suboptimal parameter settings. And training in an end-to-end fashion is still a long road.
	
	Compared with deep networks on static data, Table \ref{result1} and Table \ref{result2} demonstrate the powerful capability and popularity trend of deep spatio-temporal networks. For instance, comparison results on widely evaluated benchmarks (e.g., CK+ and MMI) illustrate that training networks based on sequence data and  analyzing temporal dependency between frames can further improve the performance. Also, in the EmotiW challenge 2015, only one system employed deep spatio-networks for FER, whereas 5 of 7 reviewed systems in the EmotiW challenge 2017 relied on such networks.
	\section{Additional Related Issues}
	\label{addition}
	In addition to the most popular basic expression classification task reviewed above, we further introduce a few related issues that depend on deep neural networks and prototypical expression-related knowledge.
	\subsection{Occlusion and non-frontal head pose}
	Occlusion and non-frontal head pose, which may change the visual appearance of the original facial expression, are two major obstacles for automatic FER, especially in real-world scenarios.
	
	For \textit{facial occlusion}, 
	Ranzato et al. \cite{susskind2011deep,mnih2013modeling} proposed a deep generative model that used mPoT \cite{mnih2010generating} as the first layer of DBNs to model pixel-level representations and then trained DBNs to fit an appropriate distribution to its inputs. Thus, the occluded pixels in images  could be filled in by reconstructing the top layer representation using the sequence of conditional distributions.
	Cheng et al. \cite{cheng2014deep} employed multilayer RBMs with a pre-training and fine-tuning process on Gabor features to compress features from the occluded facial parts.
	Xu et al. \cite{xu2015facial} concatenated high-level learned features transferred from two CNNs with the same structure but pre-trained on different data: the original MSRA-CFW database and the MSRA-CFW database with additive occluded samples. 
	
	For \textit{multi-view FER}, Zhang et al. \cite{zhang2016deep} introduced a projection layer into the CNN that learned discriminative facial features by weighting different facial landmark points within 2D SIFT feature matrices without requiring facial pose estimation. 
	Liu et al. \cite{liu2018multi} proposed a multi-channel pose-aware CNN (MPCNN) that contains three cascaded parts (multi-channel feature extraction, jointly multi-scale feature fusion and the pose-aware recognition) to predict expression labels by minimizing the conditional joint loss of pose and expression recognition.
	Besides, the technology of generative adversarial network (GAN) has been employed in \cite{lai2018emotion, zhang2018joint} to generate facial images with different expressions under  arbitrary poses for multi-view FER.
	
	\subsection{FER on infrared data}
	Although RBG or gray data are the current standard in deep FER, these data are vulnerable to ambient lighting conditions. While, infrared images that record the skin temporal distribution produced by emotions are not sensitive to illumination variations, which may be
	a promising alternative for investigation of facial expression.
	For example, He et al. \cite{he2013facial} employed a DBM model that consists of a Gaussian-binary RBM and a binary RBM for FER. The model was trained by layerwise pre-training and joint training and was then fine-tuned on long-wavelength thermal infrared images to learn thermal features. 
	Wu et al. \cite{wu2017nirexpnet} proposed a three-stream 3D CNN to fuse local and global spatio-temporal features on illumination-invariant near-infrared images for FER.

	\subsection{FER on 3D static and dynamic data}
	Despite significant advances have achieved in 2D FER, it fails to solve the two main problems: illumination changes and pose variations \cite{pantic2000automatic}. 3D FER that uses 3D face shape models with depth information can capture subtle facial deformations, which are naturally robust to pose and lighting variations. 
	
	Depth images and videos record the intensity of facial pixels based on distance from a depth camera, which contain critical information of facial geometric relations. For example, \cite{ijjina2014facial} used kinect depth sensor to obtain gradient direction information and then employed CNN on unregistered facial depth images for FER. 
	\cite{uddin2017facial, uddin2017facial1} extracted a series of salient features from depth videos and combined them with deep networks (i.e., CNN and DBN) for FER. 
	To emphasize the dynamic deformation patterns of facial expression motions,  Li et al. \cite{li2018automatic} explore the 4D FER (3D FER using dynamic data) using a dynamic geometrical image network.
	Furthermore, Chang et al. \cite{chang2018expnet} proposed to estimate
	3D expression coefficients from image intensities using CNN without requiring  facial landmark detection. Thus, the model is highly robust to extreme appearance variations, including out-of-plane head rotations, scale changes, and occlusions.
	
	Recently, more and more works  trend to combine 2D and 3D data to further improve the performance. Oyedotun et al. \cite{oyedotun2017facial} employed CNN to jointly learn facial expression features from both RGB and depth map latent modalities. 
	And Li et al. \cite{li2017multimodal} proposed  a deep fusion CNN (DF-CNN) to explore multi-modal 2D+3D FER. Specifically, six types of 2D facial attribute maps (i.e., geometry, texture, curvature, normal components \textit{x}, \textit{y}, and \textit{z}) were first extracted from the textured 3D face scans and were then jointly fed into the feature extraction and feature fusion subnets to learn the optimal combination weights of 2D and 3D facial representations. To improve this work, 
	\cite{jan2018accurate} proposed to extract deep features from different facial parts extracted from the texture and depth images, and then fused these features together to interconnect them with feedback.
	Wei et al. \cite{wei2018unsupervised} further explored the data bias problem in 2D+3D FER using unsupervised domain adaption technique.

	\subsection{Facial expression synthesis}
	Realistic \textit{facial expression synthesis}, which can generate various facial expressions for interactive interfaces, is a hot topic. Susskind et al. \cite{susskind2008generating} demonstrated that DBN has the capacity to capture the large range of variation in expressive appearance and can be trained on large but sparsely labeled datasets. In light of this work, 
	\cite{sabzevari2010fast,susskind2011deep,mnih2013modeling} employed DBN with unsupervised learning to construct facial expression synthesis systems.
	Kaneko et al. \cite{kaneko2016adaptive} proposed a multitask deep network with state recognition and key-point localization to adaptively generate visual feedback to improve  facial expression recognition. 
	With the  recent  success  of  the  deep generative models, such as variational autoencoder (VAE), adversarial autoencoder (AAE), and generative  adversarial network (GAN), a series of facial expression synthesis systems have been developed based on these models (e.g., \cite{yeh2016semantic}, \cite{zhou2017photorealistic}, \cite{song2017geometry},
	\cite{ding2017exprgan} and \cite{qiao2018geometry}).
	Facial expression synthesis can also be applied to data augmentation without manually collecting and labeling huge datasets. Masi et al. \cite{masi2016we} employed CNN to synthesize new face images by increasing  face-specific appearance variation, such as expressions within the 3D textured face model.
	
	\subsection{Visualization techniques}
	In addition to utilizing CNN for FER, several works (e.g., \cite{khorrami2015deep,mousavi2016understanding,breuer2017deep}) employed \textit{visualization} techniques \cite{zeiler2014visualizing} on the learned CNN features to qualitatively analyze how the CNN contributes to the appearance-based learning process of FER and to qualitatively decipher which portions of the face yield the most discriminative information. The deconvolutional results all indicated that the activations of some particular filters on the learned features have strong correlations with the face regions that correspond to facial AUs.
	
	\subsection{Other special issues}
	Several novel issues have been approached on the basis of the prototypical expression categories: dominant and complementary emotion recognition challenge \cite{lusi2017joint} and the Real versus Fake expressed emotions challenge \cite{wan2017results}. Furthermore, deep learning techniques have been thoroughly applied by the participants of these two challenges (e.g., \cite{kim2017discrimination,li2017combining,guo2017multi}).
	Additional related real-world applications, such as the Real-time FER App for smartphones \cite{song2014deep,bazrafkan2017deep}, Eyemotion (FER using eye-tracking cameras) \cite{hickson2017eyemotion}, privacy-preserving mobile analytics \cite{ossia2017hybrid}, Unfelt emotions \cite{kulkarni2017automatic} and Depression recognition \cite{8344107}, have also been developed.
	
	\section{Challenges and Opportunities}
	\label{further}
	\subsection{Facial expression datasets}
	As the FER literature shifts its main focus to the challenging in-the-wild environmental conditions, many researchers have committed to employing deep learning technologies to handle difficulties, such as illumination variation, occlusions, non-frontal head poses, identity bias and the recognition of low-intensity expressions. Given that FER is a data-driven task and that training a sufficiently deep network to capture subtle expression-related deformations requires a large amount of training data, the major challenge that deep FER systems face is the lack of training data in terms of both quantity and quality.
	
	Because people of different age ranges, cultures and genders display and interpret facial expression in different ways, an ideal facial expression dataset is expected to include abundant sample images with precise face attribute labels, not just expression but other attributes such as age, gender and ethnicity, which would facilitate related research on cross-age range, cross-gender and cross-cultural FER using deep learning techniques, such as multitask deep networks and transfer learning. In addition, although occlusion and multipose problems have received  relatively wide interest in the field of deep face recognition, the occlusion-robust and pose-invariant issues have receive less attention in deep FER. One of the main reasons is the lack of a large-scale facial expression dataset with occlusion type and head-pose annotations. 
	
	On the other hand, accurately annotating a large volume of image data with the large variation and complexity of natural scenarios is an obvious impediment to the construction of expression datasets. A reasonable approach is to employ crowd-sourcing models \cite{li2017reliable,Mollahosseini2017AffectNet,barsoum2016training} under the guidance of expert annotators. Additionally, a fully automatic labeling tool \cite{benitez2016emotionet} refined by experts is alternative to provide approximate but efficient annotations. In both cases, a subsequent reliable estimation or labeling learning process is necessary to filter out noisy annotations. In particular, few comparatively large-scale datasets that consider real-world scenarios and contain a wide range of facial expressions have recently become publicly available, i.e., EmotioNet \cite{benitez2016emotionet}, RAF-DB \cite{li2017reliable,li2018reliable} and AffectNet \cite{Mollahosseini2017AffectNet}, and we anticipate that with advances in technology and the wide spread of the Internet, more complementary facial expression datasets will be
	constructed to promote the development of deep FER.
	
	\subsection{Incorporating other affective models}
	Another major issue that requires consideration is that while FER within the categorical model is widely acknowledged and researched, the definition of the prototypical expressions covers only a small portion of specific categories and cannot capture the full repertoire of expressive behaviors for realistic interactions. Two additional models were developed to describe a larger range of emotional landscape: the FACS model \cite{ekman2002facial,ekman1997face}, where various facial muscle AUs are combined to describe the visible appearance changes of facial expressions,
	and the dimensional model \cite{gunes2013categorical,russell1980circumplex}, where two continuous-valued variables, namely, valence  and arousal, are proposed to continuously encode small changes in the intensity of emotions. Another novel definition, i.e., compound expression, was proposed by Du et al. \cite{du2014compound}, who argued that some facial expressions are actually combinations of more than one basic emotion. These works improve the  characterization of facial expressions and, to some extent, can complement the categorical model. For instance, as discussed above, the visualization results  of CNNs have demonstrated a
	certain congruity between the learned representations and the facial areas defined by AUs. Thus, we can design filters of the deep neural networks to distribute different weights according to the importance degree of different facial muscle action parts.
	
	\subsection{Dataset bias and imbalanced distribution}
	Data bias and inconsistent annotations are very common among different facial expression datasets due to different collecting conditions and the subjectiveness of annotating. Researchers commonly evaluate their algorithms within a specific dataset and can achieve satisfactory performance. However, early cross-database experiments have indicated that discrepancies between databases exist due to the different collection environments and construction indicators \cite{shan2009facial}; hence, algorithms evaluated via intra-database protocols lack generalizability on unseen test data, and the performance in cross-dataset settings is greatly deteriorated. Deep domain adaption and knowledge distillation  are alternatives to address this bias \cite{wei2018unsupervised,li2018domain}. Furthermore, because of the inconsistent expression annotations, FER performance cannot keep improving when enlarging the training data by directly merging multiple datasets \cite{zeng2018facial}. 
	
	Another common problem in facial expression is class imbalance, which is a result of the practicalities of data acquisition: eliciting and annotating a smile is easy, however, capturing  information for disgust, anger and other less common expressions can be very challenging. As shown in Table \ref{result1} and Table \ref{result2}, the performance assessed in terms of mean accuracy, which assigns equal weights to all classes, decreases when compared with the accuracy criterion, and this decline is especially evident in real-world datasets (e.g., SFEW 2.0 and AFEW). One solution is to balance the class distribution during the pre-processing stage using data augmentation and synthesis. Another alternative is to develop a cost-sensitive loss layer for deep networks during training.
	
	\subsection{Multimodal affect recognition}
	Last but not the least, human expressive behaviors in realistic applications involve encoding from different perspectives, and the facial expression is only one modality.
	Although pure expression recognition based on visible face images can achieve promising results, incorporating with other models into a high-level framework can provide complementary information and further enhance the robustness. For example, participants in the EmotiW challenges and Audio Video Emotion Challenges (AVEC) \cite{valstar2016avec,ringeval2017avec} considered the audio model to be the second most important element and employed various fusion techniques for multimodal affect recognition. Additionally, the fusion of other modalities, such as infrared images, depth information from 3D face models and physiological data, is becoming a promising research direction due to the large complementarity for facial expressions.

	\ifCLASSOPTIONcaptionsoff
	\newpage
	\fi

	\bibliographystyle{IEEEtran}
	\bibliography{IEEEabrv,my.bib}

\end{document}